%% file: main.tex
\documentclass[lettersize,journal]{IEEEtran}
\usepackage{amsmath,amsfonts}
\usepackage{array}
\usepackage{textcomp}
\usepackage{subfigure}
\usepackage{url}
\usepackage{verbatim}
\usepackage{graphicx}
\usepackage{cite}
\usepackage{balance}

\usepackage{amsmath,amssymb} % define this before the line numbering.
\usepackage{microtype}
\usepackage{multirow}
\usepackage{diagbox}
\usepackage{makecell}
\usepackage{dsfont}
\usepackage{hyperref}       % hyperlinks
\usepackage{url}            % simple URL typesetting
\usepackage{booktabs}       % professional-quality tables
\usepackage{amsfonts}       % blackboard math symbols
\usepackage{nicefrac}       % compact symbols for 1/2, etc.
\usepackage{microtype}      % microtypography
\usepackage{xcolor}         % colors
\usepackage{amsthm}
\usepackage[linesnumbered, lined, boxed, commentsnumbered, ruled, vlined]{algorithm2e}

\newcommand{\eat}[1]{}

\newcommand{\rev}[1]{{#1}}
\newcommand{\revsec}[1]{{#1}}
\newcommand{\revfin}[1]{{#1}}
\newcommand{\revfour}[1]{{#1}}

\newcommand{\smalltitle}[1]{ \vspace{1mm}{\noindent\textbf{#1.}\hspace{1mm}}}

\newtheorem{lemma}{Lemma}
\newtheorem{theorem}{Theorem}
\newtheorem{definition}{Definition}

\def\S{\mathcal{S}}
\def\F{\mathcal{F}}
\def\N{\mathcal{N}}
\def\E{\mathcal{E}}
\def\B{\mathcal{B}}
\def\R{\mathcal{R}}
\DeclareMathOperator*{\argmin}{arg\,min}

\begin{document}

\title{Training-Free Restoration of Pruned Neural Networks}

\author{Keonho Lee, Minsoo Kim, and Dong-Wan Choi
\thanks{K. Lee is with the Hyundai Motors, Korea. E-mail:keonho.lee@hyundai.com; M. Kim is with the KoreaPDS, Korea. E-mail:pos02043@gmail.com; D.W. Choi is with the Department of Computer Science and Engineering, Inha University, Korea. E-mail:dchoi@inha.ac.kr.}%
\thanks{D.W. Choi is the corresponding author.}
% \thanks{Manuscript received April 19, 2021; revised August 16, 2021.}
}

% The paper headers
% \markboth{Journal of \LaTeX\ Class Files,~Vol.~14, No.~8, August~2021}%
% {Shell \MakeLowercase{\textit{et al.}}: A Sample Article Using IEEEtran.cls for IEEE Journals}

% \IEEEpubid{0000--0000/00\$00.00~\copyright~2021 IEEE}
% Remember, if you use this you must call \IEEEpubidadjcol in the second
% column for its text to clear the IEEEpubid mark.

\maketitle
\input{abstract}
\input{intro}

\input{related}
\input{problem}
\input{method}

\input{experiments}

\input{conclusion}
\bibliography{references}   % name your BibTeX data base
\bibliographystyle{IEEEtran}
\balance

\input{appendix}

\vfill

\end{document}

%% file: abstract.tex
\begin{abstract}
Although network pruning has been highly popularized to compress deep neural networks, its resulting accuracy heavily depends on a fine-tuning process that is often computationally expensive and requires the original data. However, this may not be the case in real-world scenarios, and hence a few recent works attempt to restore pruned networks without any expensive retraining process. Their strong assumption is that every neuron being pruned can be replaced with another one quite similar to it, but unfortunately this does not hold in many neural networks, where the similarity between neurons is extremely low in some layers. In this article, we propose a more rigorous and robust method of restoring pruned networks in a fine-tuning free and data-free manner, called \textit{LBYL} (\textit{Leave Before You Leave}). LBYL significantly relaxes the aforementioned assumption in a way that each pruned neuron \textit{leaves} its pieces of information to as many preserved neurons as possible and thereby multiple neurons together obtain a more robust approximation to the original output of the neuron who just \textit{left}. Our method is based on a theoretical analysis on how to formulate the reconstruction error between the original network and its approximation, which nicely leads to a closed form solution for our derived loss function. Through the extensive experiments, LBYL is confirmed to be indeed more effective to approximate the original network and consequently able to achieve higher accuracy for restored networks, compared to the recent approaches exploiting the similarity between two neurons. The very first version of this work, which contains major technical and theoretical components, was submitted to NeurIPS 2021 and ICML 2022.
\end{abstract}


%% file: intro.tex
\section{Introduction} \label{sec:intro}

% 1. Network pruning has been extensively studied in neural networks. 
%    its simplicity, reasonable performance
Network pruning is one of the most well-studied model compression technique for an overparameterized deep neural network, and many different pruning strategies have been introduced with the goal of removing less important parameters from the original network. Due to its simplicity in terms of both implementation and methodology, network pruning has been pretty popular to the point that it is provided as a default model compression function in standard libraries like TensorFlow\footnote{https://www.tensorflow.org/model\_optimization/guide/pruning} and PyTorch\footnote{https://pytorch.org/tutorials/intermediate/pruning\_tutorial.html}, along with the other compression options such as quantization \cite{Coreset}. Among two major pruning schemes, namely weight pruning (\textit{a.k.a.} unstructured pruning) \cite{Lottery,Songhan,SNIP,Synaptic,OptimalPruning,SparseNet} and filter pruning (\textit{a.k.a.} structured pruning), filter pruning is more actively studied \cite{FPGM,Lasso,EagleEye,Rethinking,Thinet,DCP} due to its strength of reducing the actual computation cost at inference time without any special hardware support. In this sense, this article also focuses on filter pruning.

% \cite{GlobalRanking,AMC,FPGM,Lasso,EagleEye,Dynamic,Rethinking,CURL,Thinet,Importance,Reborn,NISP,DCP}

% To deploy large neural networks to resource-constrained mobile and edge devices, network pruning has been extensively studied due to implementation-friendly and compatibility with other network compression methods \cite{DBLP:conf/eccv/HeLLWLH18, DBLP:conf/cvpr/HeLWHY19} In general, Network pruning methods is fall in to two parts : Weight pruning(Unstructured pruning) and Filter pruning(Structured pruning). Weight pruning encourage parameters of neural network sparse during the training process or post process. As a result, Number of parameters in neural networks can be reduced parameters more than 90\% \cite{NIPS2015_ae0eb3ee} However, Because of irregular sparse connections Weight pruning need a special software to accelerate inference speed. On the other hand, Filter pruning removes unimportant filter so can accelerate inference time without special software. 

% 2. Most of pruning methods require an expensive training process
%    e.g., iterative training(on the fly), fine-tuning(post process)
% 3. A long fine-tuning eventually determines the final accuracy 
%    That is why existing pruning methods use only a few 2-3 epochs of fine-tuning

Even though many of the existing pruning methods achieve a reasonable level of the final accuracy, it mostly comes at an expensive cost of retraining at the end of (or in the middle of) pruning phase. For instance, iterative filter pruning methods \cite{Soft,FPGM,Dynamic} alternately perform pruning phase and retraining phase so that the original model can gradually be compressed as well as recovered over the hundreds of epochs. Another major approach is to conduct \textit{one-shot pruning} and then fine-tune the pruned network for its original performance to be recovered \cite{EagleEye,CURL,Importance,NISP,DCP}. In most one-shot pruning techniques, the fine-tuning phase does not take longer than a few dozens of epochs probably because there is no big difference in accuracy among pruning criteria beyond that point. In fact, it is reported that even a randomly pruned network can reach comparable performance after fine-tuning of hundreds or thousands of epochs \cite{Rethinking}. Thus, if we do not have any constraints on time, computing resources, or training data for a long fine-tuning process, one-shot pruning methods would be less meaningful.

% 4. Therefore, they obviously need a quiet complete set of training data
%    (mostly exactly the same as the original train set) or need to collect as 
%    many samples as the origin
% 5. However, in practice, the train data is neither available nor easy to collect
% 6. Therefore, a few recent works have tried to address above with using much small data ( still need data, train process anyway)
% 7. Meanwhile, a few works commonly employ the neuron similarity in order to 
%    reduce the bad effect of pruning Thus, a similar neuron can be replaced
%    limitations of NM, 
%    1) it is based on strong naive assumption, which is not the case in NNs.
%    2) a single neuron can't compensate due to the error term we found yet not discussed

\begin{figure*}[t!]
   \centering 
    \includegraphics[width=1.9\columnwidth]{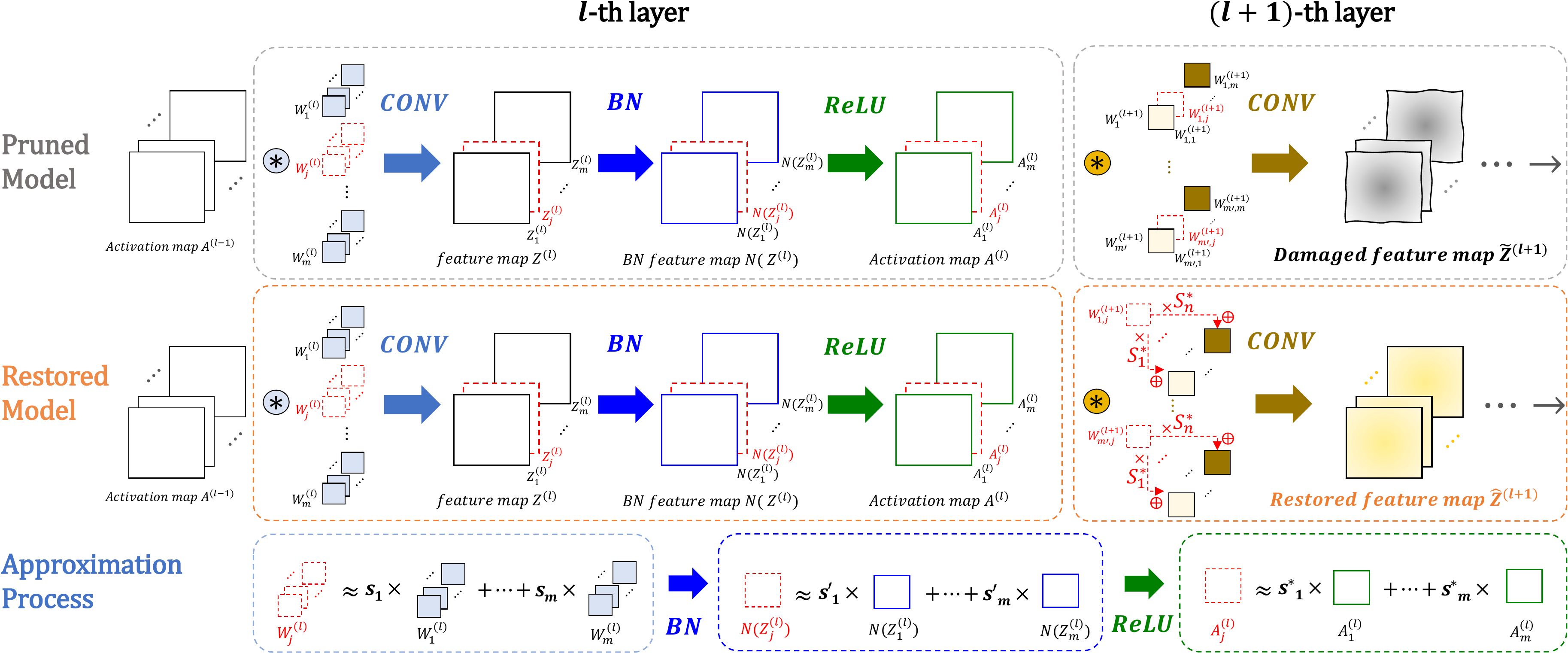}
   \caption{The conceptual overview of our LBYL method, showing how the original output resulting from a pruned filter at $\ell$-th layer, that is, the output of $(\ell+1)$-th convolutional layer, can be recovered by all the other preserved filters at the same layer (\textit{i.e.}, $\ell$-th layer) through convolution, batch normalization, and activation function (\textit{i.e.}, ReLU), where $s, s',$ and $s^*$ are the coefficients that quantify how much each preserved filter should carry the information of the pruned filter.}
   \label{fig:framework}
\end{figure*}

Somewhat fortunately, however, this is not the case in many practical scenarios where we cannot access or re-collect the original data due to its large volume or some commercial issues \cite{DBLP:conf/eccv/MahajanGRHPLBM18}. Even though some recent works \cite{CURL,Reborn} attempt to use only a small amount of training data instead, but they still suffer from a time-consuming process of fine-tuning to achieve satisfactory performance. Furthermore, it would not be always possible to collect samples in the same domain as the original data, regardless of whether they are labeled or not. Only quite a few works \cite{NM,Data-free} have been introduced to recover pruned networks without using any data and fine-tuning process. Their common approach is what we call \textit{one-to-one compensation}, where the most similar unpruned neuron takes the role of a pruned neuron. This approach is based on the strong assumption that there exists such a neuron quite similar to each one being pruned at the same layer. Not surprisingly, this assumption cannot be guaranteed in deep neural networks, which is even reported in the result of \cite{NM} showing that the pairwise similarity on filters gets extremely low particularly in deeper layers. In addition, according to our theoretical analysis on the reconstruction error, a single neuron can only lead to a very rough approximation to the original output even if the neuron is similar enough to its pruned counterpart.

In this article, we propose a theoretically more rigorous and robust method, called \textit{Leave Before You Leave} (\textit{LBYL}), to effectively compensate for pruned filters, which is also free of any data and fine-tuning. As depicted in Figure \ref{fig:framework}, we significantly relax the strong assumption by observing the fact that a pruned filter is better to be represented by a linear combination of as many preserved filters as possible than it is done by a single filter. We first mathematically analyze how the reconstruction error can be formulated between the original network and its pruned network, and thereby discover three components of the error. Based on our formulation, we provide a closed form solution that can minimize the reconstruction error to compensate for information loss caused by pruned filters, which does not need any training process at all. Through extensive experiments, we show that LBYL outperforms the existing one-to-one compensation scheme \cite{NM} regardless of pruning criterion. For instance, LBYL achieves at pruning ratio 30\% on the average 17.92\% higher accuracy than those of \cite{NM} for ResNet-101 \cite{ResNet} trained with ImageNet \cite{deng2009imagenet}.

% LBYL achieves at pruning ratio 40\% on average 3.53\% higher accuracy than \cite{NM} for ResNet-50 \cite{ResNet} on CIFAR-100 \cite{krizhevsky2009learning} and also in case of large dataset (i.e., ImageNet \cite{deng2009imagenet}), .

% using CIFAR \cite{krizhevsky2009learning} and ImageNet \cite{deng2009imagenet}

%% file: related.tex
\section{Related Works} \label{sec:related}
Many filter pruning methods have been studied without considering such a case where we cannot use any data, whether real or synthetic, to restore a pruned network. They mostly go through an expensive retraining process either iteratively \cite{Soft,FPGM,Dynamic,ZhangF24} or lastly as a fine-tuning phase \cite{GlobalRanking,Importance,NISP}. In this article, we focus on the opposite case, which is more challenging in practice, where fine-tuning with any data is not available.

\smalltitle{Pruning with less fine-tuning} 
Most of the existing filter pruning methods have tried to avoid an expensive fine-tuning process by means of a carefully designed criterion for identifying unimportant filters such that the loss of information is minimized when they are pruned. To this end, they often make the best use of data-dependent values like channels and gradients that can be obtained by making inferences with some data. \cite{Thinet} proposes a greedy algorithm that prunes the filters whose corresponding channels minimize the layer-wise reconstruction error. Similarly, \cite{Lasso} formulates this problem of channel selection as lasso regression and least square reconstruction. \revfour{\cite{URC} globally chooses the redundant filters in the pre-trained model by utilizing feature-discrimination-based filter importance. \cite{catro} suggests uninformative channel selection method via class-aware trace ratio optimization, which measures joint impact across channels in a convolutional layer.} \cite{CoreSet_ICLR} introduces data-independent pruning criterion based on coreset together with an intermediate recovery method without training in order to reduce the overhead of fine-tuning. Despite their proposed methods on effective channel or neuron selection, a few epochs of fine-tuning process as well as some training data is inevitable for the pruned network to be sufficiently recovered. \revfour{To speedup filter pruning, \cite{LeeS24} utilizes a finetuning structure based on constrastive knowledge transfer, thereby proposing a coarse-to-fine neural architecture search algorithm.}

\smalltitle{Pruning with less fine-tuning and less data}
For the case of pruning with limited data, several methods \cite{CURL,Reborn} utilize knowledge distillation \cite{Knowledge_Distilation} not to use the entire original data in the recovery process. \cite{CURL} recently proposes \textit{CURL} that globally prunes filters using a KL-divergence based criterion and perform knowledge distillation with a small dataset as a fine-tuning process. \cite{Reborn} tackles the similar problem and devise a way of transforming all the filters in a layer into compact ones, called \textit{reborn filters}, using only the limited data. These methods do not use the entire original data, but they still require some kinds of retraining process with a small amount of data. 

\smalltitle{Pruning with no fine-tuning and no data} In the literature of filter pruning, there are only a few works \cite{NM, Data-free,RedPlus} that adopt the same problem setting as ours, that is, pruning filters without any fine-tuning and data. \rev{It is worth mentioning that our work focuses on this \textit{pure} recovery process with respect to pruned networks without any support of extra memory and hardware, as in \cite{NM,Data-free}.} Both methods are based on the strong assumption that we can always find a pair of filters that are quite similar to each other and therefore a pruned filter can be compensated by its similar unpruned one. However, this assumption does not hold in modern convolutional neural networks (CNNs) having a number of filters in a fairly deep architecture. As even presented in \cite{NM}, the cosine similarity between each filter and its nearest one gets extremely low (\textit{e.g.}, around 0.2) in back layers to the point that it is better to do nothing for some pruned neurons that do not have sufficiently similar preserved neurons. In this article, we remedy this problem by using as many preserved filters as possible to compensate for each pruned filter, and thereby propose a more robust method that successfully relaxes the assumption made by these existing works. \rev{Meanwhile, the RED++ method \cite{RedPlus} more concentrates on how to make the best use of extra memory buffer (i.e., RAM) other than GPU in order to reduce the number of GPU computations. Therefore, it requires high-speed memory access with a specific hardware support, which makes orthogonal to our method.}

% \lee{It is worth mentioning that \cite{NM, Data-free} methods concentrate on recovery to performance drop after pruning without any soft/hardware supports in same way as we do. However, \cite{Red++} concentrates on how to compress network without performance drop. That is, \cite{Red++} needs special soft/hardware supports to speed up memory access and allocation than computation to accelerate network inference.
% } \cite{NM, Data-free}

% Both methods are based on the strong assumption that we can always find a pair of filters that are quite similar to each other and therefore a pruned filter can be compensated by its similar unpruned one. However, this assumption does not hold in modern convolutional neural networks (CNNs) having a number of filters in a fairly deep architecture. As even presented in \cite{NM}, the cosine similarity between each filter and its nearest one gets extremely low (\textit{e.g.}, around 0.2) in back layers to the point that it is better to do nothing for some pruned neurons that do not have sufficiently similar preserved neurons. In this article, we remedy this problem by using as many preserved filters as possible to compensate for each pruned filter, and thereby propose a more robust method that successfully relaxes the assumption made by these existing works.

\smalltitle{Data-free compression with heavy retraining}
\rev{
There is a large branch of data-free compression methods \cite{KEGNET,DAFL019,DFAD_BEFORE,DFAD,DeepInversion,Data-Free-NetworkPruning} that still fine-tune or retrain the compressed model with synthetically generated data but not real data. Thus, in some sense, they are not literally data-free, but free of original data. Most of these methods involves extracting a generator that can generate synthetic samples, which itself independently takes a considerable amount of training time, and rely on knowledge distillation, which thus makes them often called \textit{data-free knowledge distillation}. As in \cite{Data-Free-NetworkPruning,DeepInversion}, some of the works employ pruned networks as a student model, instead of a lightweight network randomly initialized, but still focus on retraining process with synthetic data. Note that these data-free methods still suffer from heavy recovery process for fine-tuning (or distillation) as well as training the generator, which is not our focus in this article.
}

% \lee{ \cite{RedPlus} focuses on reducing the number of computations by changing identical operations to identical memory access to a unique operation. That is, \cite{RedPlus} needs the hypothesis that memory access and allocation should run faster than mathematical operations, while our LBYL method do not need a specific hardware assumption. Therefore \cite{RedPlus} is orthogonal to our method. In this article, we remedy problem of similarity based compensation \cite{NM, Data-free} by using as many preserved filters as possible to compensate for each pruned filter, and thereby propose a more robust method that successfully relaxes the assumption made by these existing works.}

\smalltitle{Clarification regarding overlapping work}
We have become aware of two recently published papers—one in ICCV 2023 titled ``\textit{Unified Data-Free Compression: Pruning and Quantization without Fine-Tuning}” and another in Pattern Recognition titled ``\textit{Data-Free Quantization via Mixed-Precision Compensation without Fine-Tuning}.” Both appear to share core derivations and theoretical components with an earlier version of our submission, which was not publicly accessible at the time those papers were written. We have formally reported our concerns about these similarities to the relevant committees. As we await a resolution, we are refraining from listing these papers in our references to avoid lending legitimacy to them while the plagiarism dispute remains unsettled.

%% file: problem.tex
\section{Problem Formulation} \label{sec:probdef}

This section formally defines the problem of restoring a given pruned network with only using its original pretrained CNN in a way free of data and fine-tuning.

\begin{figure*}[t]
	\centering
    \subfigure[\label{fig:matrix:a}Pruning matrix]{\hspace{6mm}\includegraphics[width=0.4\columnwidth]{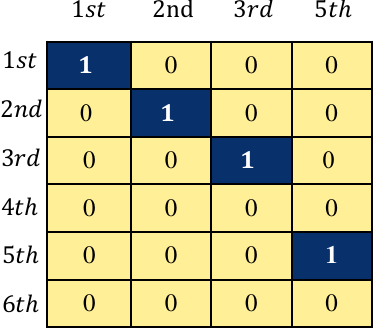}\hspace{6mm}} 
    \subfigure[\label{fig:matrix:b}Delivery matrix for LBYL]{\hspace{6mm}\includegraphics[width=0.4\columnwidth]{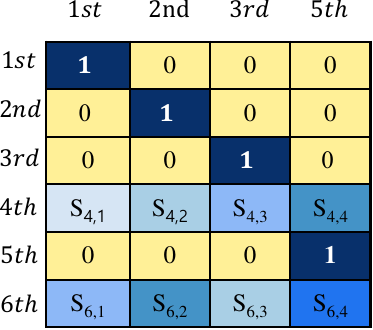}\hspace{6mm}}
    \subfigure[\label{fig:matrix:c}Delivery matrix for one-to-one]{\hspace{9mm}\includegraphics[width=0.4\columnwidth]{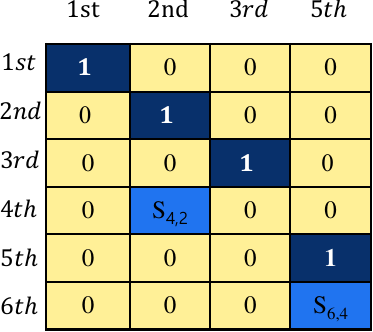}\hspace{9mm}} 
    \caption{Comparison between pruning matrix and delivery matrix, where the $4$-th and $6$-th filters are being pruned among $6$ original filters}
	\label{fig:matrix}
	\vspace{-2mm}
\end{figure*}

\subsection{Filter Pruning in a CNN}
Consider a given CNN to be pruned with $L$ layers, where each $\ell$-th layer starts with a convolution operation on its input channels, which are the output of the previous $(\ell-1)$-th layer $\mathbf{A}^{(\ell-1)}$, with the group of convolution filters $\mathbf{W}^{{(\ell)}}$ and thereby obtain the set of \textit{feature maps} $\mathbf{Z}^{(\ell)}$ as follows:
\begin{equation}
\boldsymbol{\mathbf{Z}}^{(\ell)} = {\mathbf{A}^{(\ell-1)} \circledast {\mathbf{W}}^{(\ell)}},
\nonumber
\end{equation}
where $\circledast$ represents the convolution operation. Then, this convolution process is normally followed by a batch normalization (BN) process and an activation function such as ReLU, and the $\ell$-th layer finally outputs an \textit{activation map} $\mathbf{A}^{(\ell)}$ to be sent to the $(\ell+1)$-th layer through this sequence of procedures as:
\begin{equation}
\mathbf{A}^{(\ell)} = \F(\N(\mathbf{Z}^{(\ell)})),
\nonumber
\end{equation}
where $\F(\cdot)$ is an activation function and $\N(\cdot)$ is a BN procedure.

Note that all of $\mathbf{W}^{(\ell)}$, $\mathbf{Z}^{(\ell)}$, and $\mathbf{A}^{(\ell)}$ are tensors such that: $\mathbf{W}^{(\ell)} \in \mathbb{R}^{m \times n \times k \times k}$ and $\mathbf{Z}^{(\ell)},\mathbf{A}^{(\ell)} \in \mathbb{R}^{m \times w \times h}$, where (1) $m$ is the number of filters, which also equals the number of output activation maps, (2) $n$ is the number of input activation maps resulting from the $(\ell-1)$-th layer, (3) $k \times k$ is the size of each filter, and (4) $w \times h$ is the size of each output channel for the $\ell$-th layer.

\smalltitle{Filter pruning as n-mode product}
When filter pruning is performed at the $\ell$-th layer, all three tensors above are consequently modified to their \textit{damaged} versions, namely $\mathbf{\Tilde{W}}^{(\ell)}$, $\mathbf{\Tilde{Z}}^{(\ell)}$, and $\mathbf{\Tilde{A}}^{(\ell)}$, respectively, in a way that: $\mathbf{\Tilde{W}}^{(\ell)} \in \mathbb{R}^{t \times n \times k \times k}$ and $\mathbf{\Tilde{Z}}^{(\ell)},\mathbf{\Tilde{A}}^{(\ell)} \in \mathbb{R}^{t \times w \times h}$, where $t$ is the number of remaining filters after pruning and therefore $t < m$. Mathematically, the tensor of remaining filters, \textit{i.e.}, $\mathbf{\Tilde{W}}^{(\ell)}$, is obtained by the \textit{$1$-mode product} \cite{DBLP:journals/siamrev/KoldaB09}\footnote{\revfour{Please refer to Section 2.5. in \cite{DBLP:journals/siamrev/KoldaB09}}} of the tensor of the original filters $\mathbf{W}^{(\ell)}$ with a \textit{pruning matrix} $\boldsymbol{\S} \in \mathbb{R}^{m \times t}$ (see Figure \ref{fig:matrix:a})
as follows:
\begin{eqnarray}\begin{split}\label{eq:pruning}
\mathbf{\Tilde{W}}^{(\ell)} = {\mathbf{W}}^{(\ell)} \times_{1} {\boldsymbol{\S}}^{T},\text{where }\boldsymbol{\S}_{i,k} = 
  \begin{cases} 
   1~ \text{if } i = i'_k \\
   0~ \text{otherwise}
  \end{cases} \\
  \text{s.t. } i, i'_k \in [1, m] 
  \text{ and } k \in [1, t].
  \end{split}
\end{eqnarray}
  
By Eq. (\ref{eq:pruning}), each $i'_k$-th filter is not pruned and the other $(m-t)$ filters are completely removed from $\mathbf{W}^{(\ell)}$ to be $\mathbf{\Tilde{W}}^{(\ell)}$.

This reduction at the $\ell$-th layer causes another reduction for each filter of the $(\ell+1)$-th layer so that $\mathbf{W}^{(\ell+1)}$ is now modified to $\mathbf{\Tilde{W}}^{(\ell+1)} \in \mathbb{R}^{m' \times t \times k' \times k'}$, where $m'$ is the number of filters of size $k' \times k'$ in the $(\ell+1)$-th layer. Due to this series of information losses, the resulting feature map (\textit{i.e.}, $\mathbf{Z}^{(\ell+1)}$) would severely be damaged to be $\mathbf{\Tilde{Z}}^{(\ell+1)}$ as shown below:
\begin{equation}
{\mathbf{\Tilde{Z}}}^{{(\ell+1)}} = \mathbf{\Tilde{A}}^{(\ell)} \circledast {\mathbf{\Tilde{W}}}^{(\ell+1)}~~~\not\approx~~~\mathbf{Z}^{(\ell+1)}
\label{eq:eq}\nonumber
\end{equation}
The shape of $\mathbf{\Tilde{Z}}^{(\ell+1)}$ remains the same unless we also prune filters for the $(\ell+1)$-th layer. If we do so as well, the loss of information will be accumulated and further propagated to the next layers. Note that $\mathbf{\Tilde{W}}^{(\ell+1)}$ can also be represented by the \textit{$2$-mode product} \cite{DBLP:journals/siamrev/KoldaB09} of $\mathbf{W}^{(\ell+1)}$ with the transpose of the same matrix $\boldsymbol{\S}$ as:
\begin{equation} \label{eq:pruning2}
\mathbf{\Tilde{W}}^{(\ell+1)} = {\mathbf{W}}^{(\ell+1)} \times_{2} {\boldsymbol{\S}^T}
\end{equation}

\subsection{Problem of Restoring a Pruned Network without Data and Fine-Tuning}
As mentioned earlier, our goal is to restore a pruned and thus damaged CNN without using any data and re-training process, which implies the following two facts. First, we have to use a pruning criterion exploiting only the values of filters themselves such as L1-norm. In this sense, this article does not focus on proposing a sophisticated pruning criterion but intends to recover a network somehow pruned by such a simple criterion. Secondly, since we cannot make appropriate changes in the remaining filters by fine-tuning, we should make the best use of the original network and identify how the information carried by a pruned filter can be delivered to the remaining filters.

% For brevity, we formulate our problem here with respect to a specific layer, say $\ell$, and then it can trivially be generalized for the entire network. 
\smalltitle{Delivery matrix}
In order to represent the information to be delivered to the preserved filters, let us first think of what the pruning matrix $\boldsymbol{\S}$ means. As defined in Eq. (\ref{eq:pruning}) and shown in Figure \ref{fig:matrix:a}, each row is either a zero vector (for filters being pruned) or a one-hot vector (for remaining filters), which is intended only to remove filters without delivering any information. Intuitively, we can transform this pruning matrix into a \textit{delivery matrix} that carries information for filters being pruned by replacing some meaningful values with some of the zero values therein. Once we find such an \textit{ideal} $\boldsymbol{\S^*}$, we can plug it into $\boldsymbol{\S}$ of Eq. (\ref{eq:pruning2}) to deliver missing information propagated from the $\ell$-th layer to the filters at the $(\ell+1)$-th layer, which will hopefully generate an approximation $\mathbf{\hat{Z}}^{(\ell+1)}$ close to the original feature map as follows:
\begin{equation} \label{eq:fmap_approx}
{\mathbf{\hat{Z}}}^{{(\ell+1)}} = {\mathbf{\Tilde{A}}^{(\ell)} \circledast ({\mathbf{W}}^{(\ell+1)} \times_{2} {\boldsymbol{\S^*}^T})}
~~~\approx~~~\mathbf{Z}^{(\ell+1)}
\end{equation}
Thus, using the delivery matrix $\boldsymbol{\mathcal{S^*}}$, the information loss caused by pruning at each layer is recovered at the feature map of the next layer. \revfour{It is noteworthy that here both pruning matrix and delivery matrix are theoretical concepts to see the insight on what is happening when pruning filters from a pretrained CNN, rather than things that actually exist in neural networks.}

\smalltitle{Problem statement}
Given a pretrained CNN, our problem aims to find the best delivery matrix $\boldsymbol{\mathcal{S^*}}$ for each layer without any data and training process such that the following \textit{reconstruction error} is minimized:
\begin{equation}
\sum\limits_{i = 1}^{m'}\|{{\mathbf{Z}}_{i}^{{(\ell+1)}}-{\hat{\mathbf{Z}}}_{i}^{{(\ell+1)}}}\|_1,
\label{eq:goal}
\end{equation}
where ${\mathbf{Z}}_i^{{(\ell+1)}}$ and ${\hat{\mathbf{Z}}}_i^{{(\ell+1)}}$ indicate the $i$-th original feature map and its corresponding approximation, respectively, out of $m'$ filters in the $(\ell+1)$-th layer. Note that what is challenging here is that we cannot obtain the activation maps in $\mathbf{A}^{(\ell)}$ and $\mathbf{\Tilde{A}}^{(\ell)}$ without data as they are data-dependent values.

%% file: method.tex
\section{Proposed Method of Restoring Pruned Filters} \label{sec:method}
In this section, we first theoretically examine how the reconstruction error of Eq. (\ref{eq:goal}) can be formulated in a way free of using data-dependent parameters. Based on this analysis, we present our train-free recovery method that has a closed form solution for minimizing the formulated reconstruction loss.

%is composed of with respect to our proposed methodology using multiple filters to approximate a pruned filter, so-called \textit{many-to-one}. Our approach is in fact a generalized version of the existing one-to-one compensation methods \cite{NM,Data-free}. Finally, we present our train-free recovery method that has a closed form solution for minimizing the formulated reconstruction error.

\subsection{Data-Independent Reconstruction Loss} \label{sec:method:a}
For brevity, we deal with a simple case of pruning only a single filter, which can trivially be extended to cover multiple filters. Without loss of generality, let the $j$-th filter of the $\ell$-th layer be pruned. \revfour{Thus, here after, $j$ is a fixed value that represents a particular filter being pruned.}

\smalltitle{Leave Before You Leave (LBYL) assumption}
In order to transform Eq. (\ref{eq:goal}) to a data-free version, our proposed assumption is that each filter being pruned can \textit{leave} its pieces of information for the remaining ones by adding its corresponding pruned channel to each of the other channels in the next layer. Since the quantity of information depends on how much the pruned filter is related to a particular remaining filter, we need to multiply the pruned channel with a different value before addition. Based on this assumption, \revfour{for the $j$-th filter being pruned}, the delivery matrix $\boldsymbol{\S^*} \in \mathbb{R}^{m \times (m-1)}$ is defined to be in the following form:
\begin{eqnarray}\begin{split}\label{eq:matrix}
\boldsymbol{\S^*}_{i,k} = 
  \begin{cases} 
   1 & \text{if } i = i'_k \\
   s_{i'_k} & \text{if } i = j \\
   0 & \text{otherwise}
  \end{cases}
  \end{split}
\end{eqnarray}
 $\text{s.t. } i, i'_k \in [1, m] \text{ and } k \in [1, m-1],$ where $s_{i'_k}$ is a scalar value that quantifies to what extent the $i'_k$-th filter is related to the $j$-th filter being pruned. Figure \ref{fig:matrix:b} shows an example of the delivery matrix for our LBYL method.

\smalltitle{Without BN and an activation function}
Starting with Eq. (\ref{eq:goal}), we first derive its data-free version without considering both a BN procedure and an activation function, and then make an extension for them. With the proposed form of the delivery matrix in Eq. (\ref{eq:matrix}) together with Eq. (\ref{eq:fmap_approx}), the $i$-th channel of $\mathbf{\hat{Z}}^{{(\ell+1)}}$ is now represented as:
\begin{equation}
{{\mathbf{\hat{Z}}}_{i}^{{(\ell+1)}}} = \sum\limits_{k = 1, k \neq j}^{m} \mathbf{A}_{k}^{(\ell)} \circledast (\mathbf{W}_{i,k}^{(\ell+1)}+{s_{k}} \mathbf{W}_{i,j}^{(\ell+1)}).
\nonumber
\end{equation}
Then, the reconstruction error for the $i$-th channel of the feature map at the $(\ell+1)$-th layer can be derived as:
\begin{equation}
{\mathbf{Z}_{i}^{{(\ell+1)}}}-{\mathbf{\hat{Z}}_{i}^{{(\ell+1)}}}=(\mathbf{A}_{j}^{(\ell)}-\sum\limits_{k = 1, k \neq j}^{m}{s_{k}} \mathbf{A}_{k}^{(\ell)})\circledast \mathbf{W}_{i,j}^{(\ell+1)}.
\label{eq:eq5}
\end{equation}
By Eq. (\ref{eq:eq5}), we have the following form of the reconstruction error for the case of pruning the $j$-th filter at the $\ell$-th layer:
\begin{eqnarray}\begin{split}
\sum\limits_{{i} = 1}^{m^{\prime}}\|{\mathbf{Z}_{i}^{{(\ell+1)}}-\mathbf{\hat{Z}}_{i}^{{(\ell+1)}}}\|_1 ~~~~~~~~~~~~~~~~~\\ 
=\sum\limits_{{i} =  1}^{m^{\prime}}\|(\mathbf{A}_{j}^{(\ell)}-\sum\limits_{k = 1, k \neq j}^{m}{s_{k}} \mathbf{A}_{k}^{(\ell)})\circledast\mathbf{W}_{i,j}^{(\ell+1)}\|_1.
\end{split}
\label{eq:eq6}
\end{eqnarray}
Based on Eq. (\ref{eq:eq6}), we can reduce the problem of minimizing Eq. (\ref{eq:goal}) into the one of minimizing the following error term:
\begin{equation} \label{eq:goal2}
\|(\mathbf{A}_{j}^{(\ell)}-\sum\limits_{k = 1, k \neq j}^{m}{s_{k}} \mathbf{A}_{k}^{(\ell)})\|_1,
\end{equation}
where we need to find scalars $s_{k}$'s such that ${ \mathbf{A}_{j}^{(\ell)} \approx \sum\limits_{k = 1, k \neq j}^{m}{s_{k}} \mathbf{A}_{k}^{(\ell)}}$. This is because the term $\mathbf{W}_{i,j}^{(\ell+1)}$ in Eq. (\ref{eq:eq6}) is a constant with respect to a given pretrained model. Therefore, minimizing Eq. (\ref{eq:goal2}) makes the same effect as minimizing Eq. (\ref{eq:eq6}). \revfour{More specifically, this is based on the fact that the convolution operation $\mathbf{Z} = \mathbf{A} \circledast \mathbf{W}$ is linear with respect to both participating terms $\mathbf{A}$ and $\mathbf{W}$. Therefore, if $\mathbf{W}$ is constant, reducing the values in $\mathbf{A}$ will directly reduce the output values $\mathbf{Z}$ resulting from the convolution operation, as $\mathbf{Z}$ is directly proportional to $\mathbf{A}$ for fixed $\mathbf{W}$.} Without considering BN and any activation function, we simply have $\mathbf{A}^{(\ell)} = \mathbf{Z}^{(\ell)}$, and consequently Eq. (\ref{eq:goal2}) can be represented as:
\begin{gather*}
    %  \|\mathbf{A}_{j}^{(\ell)}-\sum\limits_{k = 1, k \neq j}^{m}{s_{k}} \mathbf{A}_{k}^{(\ell)}\|_{1} \\
    % = \| (\mathbf{A}^{(\ell-1)}  \circledast \mathbf{W}_{j}^{(\ell)} ) - \sum\limits_{k = 1, k \neq j }^{m} {s}_{k} \times (\mathbf{A}^{(\ell-1)}  \circledast \mathbf{W}_{k}^{(\ell)} ) \|_{1} \nonumber \\
    % = 
     \| (\mathbf{A}^{(\ell-1)}  \circledast (\mathbf{W}_{j}^{(\ell)} - \sum\limits_{k = 1, k \neq j }^{m} {s}_{k} \times \mathbf{W}_{k}^{(\ell)} ) \|_{1}.
\end{gather*}
Since $\mathbf{A}^{(\ell-1)}$ is data-dependent yet independent to pruned filters, it can be regarded as a constant factor of the reconstruction error that depends on which filters are pruned. Thus, it suffices to minimize the other part of the term, which we call \textit{Residual Error} (\textit{RE}) denoted by $\boldsymbol{\E}$ as the following definition:
\begin{definition} \label{def:re}
Given a pruned filter $\mathbf{W}_{j}^{(\ell)}$ at the $\ell$-th layer, the Residual Error (RE) is defined as:
$$
\boldsymbol{\E} = \mathbf{W}_{j}^{(\ell)} - \sum\limits_{k = 1, k \neq j }^{m} {s}_{k} \times \mathbf{W}_{k}^{(\ell)}.
$$
\end{definition}
Note that $\boldsymbol{\E}$ can be minimized by using only the original network without any support from data.

% Since we cannot access $\mathbf{A}^{(\ell-1)}$ without data, the only thing we can do is minimizing the other part of the term. We call this error term \textit{Residual Error} (\textit{RE}) and denote it by $\boldsymbol{\E}$

\smalltitle{Considering batch normalization} 
We now incorporate a BN procedure in the process of minimizing Eq. (\ref{eq:goal2}). In this case, as we have $\mathbf{A}^{(\ell)} = \N(\mathbf{Z}^{(\ell)})$, Eq. (\ref{eq:goal2}) can be derived as Lemma \ref{lem:bn}.
\begin{lemma} \label{lem:bn}
If there is only batch normalization between a feature map and its activation map, the reconstruction error can be formulated as follows:
\begin{equation} \label{eq:bn}
\| \frac{\gamma_{j}}{\sigma_{j}}{(\mathbf{A}^{(\ell-1)} \circledast \boldsymbol{\E})} + \boldsymbol{\B}\|_{1},
\end{equation}
% \begin{equation} \label{eq:bn}
% \| \frac{\gamma_{j}}{\sigma_{j}}{(\mathbf{A}^{(\ell-1)} \circledast \boldsymbol{\E})} +
% \frac{\gamma_{j}}{\sigma_{j}} 
% \{\sum\limits_{k = 1, k \neq j }^{m} s_{k}\frac{\sigma_{j}}{\gamma_{j}}\frac{\gamma_{k}}{\sigma_{k}} (\mu_{k} -  \frac{\sigma_{k}}{\gamma_{k}}\beta_k) - \mu_j + \frac{\sigma_{j}}{\gamma_{j}}\beta_j\} \|_{1},
% \end{equation}
where $\boldsymbol{\B} = \frac{\gamma_{j}}{\sigma_{j}} 
 \{\sum\limits_{k = 1, k \neq j }^{m} s_{k}\frac{\sigma_{j}}{\gamma_{j}}\frac{\gamma_{k}}{\sigma_{k}} (\mu_{k} -  \frac{\sigma_{k}}{\gamma_{k}}\beta_k) - \mu_j + \frac{\sigma_{j}}{\gamma_{j}}\beta_j\}$, $\mu$, $\gamma$, $\sigma$, and $\beta$ are batch normalization parameters.
\end{lemma}
\begin{proof}
See Appendix.
\end{proof}
Hereafter, the notation $\boldsymbol{\E}$ is a bit differently defined such that $s_k$ from Definition \ref{def:re} should be substituted with $s_k \frac{\sigma_j}{\gamma_j} \frac{\gamma_k}{\sigma_k}$ due to the BN procedure. We call the second term $\boldsymbol{\B}$ of Eq. (\ref{eq:bn}) Batch Normalization Error (BE).

\begin{definition}
Given a pruned filter $\mathbf{W}_{j}^{(\ell)}$ at the $\ell$-th layer, the Batch Normalization Error (BE) is defined as:
$$
\boldsymbol{\B} = \frac{\gamma_{j}}{\sigma_{j}} 
\{\sum\limits_{k = 1, k \neq j }^{m} s_{k}\frac{\sigma_{j}}{\gamma_{j}}\frac{\gamma_{k}}{\sigma_{k}} (\mu_{k} -  \frac{\sigma_{k}}{\gamma_{k}}\beta_k) - \mu_j  + \frac{\sigma_{j}}{\gamma_{j}}\beta_j \}.
$$
\end{definition}
\revsec{It is noteworthy that BN parameters are obtained through the pretraining process just like the other learned weights, and stored in the pretrained model to be used as a linear transform layer at inference time.}\footnote{https://en.wikipedia.org/wiki/Batch\_normalization}

% below Eq. (\ref{eq:eq9}).
% \begin{eqnarray} \begin{split}\label{eq:eq9}
%      \|\mathbf{A}_{j}^{(\ell)}-\sum\limits_{k = 1, k \neq j}^{m}{s_{k}} \mathbf{A}_{k}^{(\ell)}\|_{1} = &~\| \N(\mathbf{Z}^{(\ell-1)}  \circledast \mathbf{W}_{j}^{(l)} ) - \sum\limits_{k = 1, k \neq j }^{m} {s}_{k} \times \N (\mathbf{Z}^{(\ell-1)}  \circledast \mathbf{W}_{k}^{(l)} ) \|_{1}
% \end{split}
% \end{eqnarray}
% Eq. (\ref{eq:eq9}) is further derived and finally we can conclude to the following lemma:

\smalltitle{Considering activation function}
Lastly, we proceed our analysis to cover a general CNN architecture including both BN and the ReLU activation function. For an activation function $\F$, we now have $\mathbf{A}^{(\ell)} = \F(\N(\mathbf{Z}^{(\ell)}))$, and thereby Eq. (\ref{eq:goal2}) is derived to be the following theorem:
\begin{theorem} \label{thm:relu}
If there are both batch normalization and a ReLU function between a feature map and its activation map, the reconstruction error can be formulated as follows:
    \begin{equation} \label{eq:relu}
     \| \frac{\gamma_{j}}{\sigma_{j}}{(\mathbf{A}^{(\ell-1)}\circledast\boldsymbol{\E})}+ \boldsymbol{\B} + \boldsymbol{\R}\|_{1},
     \end{equation}
     where $\boldsymbol{\R} = \sum\limits_{k=1, k \neq j}^{m} s_{k}  \min(0,\N(\mathbf{Z}_{k}^{(\ell)}))-\min(0,\N(\mathbf{Z}_{j}^{(\ell)}))$ represents the \textit{Activation Error} (\textit{AE}).
\end{theorem}
\begin{proof}
See Appendix.
\end{proof}

\smalltitle{Data-free loss function}
\rev{As concluded in Theorem \ref{thm:relu}, we discover the fact that the final reconstruction error of Eq. (\ref{eq:goal}) consists of three different components, namely RE, BE, and AE.} \revsec{Since it is not feasible to find a training-free solution for minimizing Eq. (\ref{eq:relu}) itself, we focus on the fact that Eq. (\ref{eq:relu}) is smaller than equal to:
\begin{eqnarray}
         \| \frac{\gamma_{j}}{\sigma_{j}}{(\mathbf{A}^{(\ell-1)}\circledast\boldsymbol{\E})}\|_{1} + \|\boldsymbol{\B}\|_{1} + \|\boldsymbol{\R}\|_{1}. \nonumber
\end{eqnarray}
This further brings us to formulate the following reconstruction loss that does not require any training data and nicely has a closed form solution as derived in Theorem \ref{thm:closedform}:
\begin{equation}
     \mathcal{L}_{re} =  \|\boldsymbol{\E}\|_2^{2} + \lambda_1 \|\boldsymbol{\B}\|_2^{2} + \lambda_2 \|\mathbf{s}\|_{2}^{2},
\label{eq:loss}
\end{equation}
where $\mathbf{s} =  [s_{1}~...~s_{j-1}~s_{j+1}~...~s_{m}]$.
}

Unlike RE \revsec{(\textit{i.e.,} $\boldsymbol{\E}$)} and BE \revsec{(\textit{i.e.,} $\boldsymbol{\B}$)}, we cannot exactly control the AE term \revsec{(\textit{i.e.,} $\boldsymbol{\R}$)} as $\mathbf{Z}^{(\ell)}$ can only be obtained by using the actual data. Therefore, we instead introduce a regularization term $\|\mathbf{s}\|_{2}^{2}$ for the purpose of mitigating the explosion of $\boldsymbol{\R}$, \revsec{inspired by the fact that the upper bound of $\|\boldsymbol{\R}\|_{1}$ is proportional to $\sum_{k=1,k\neq j}^{m}\|s_k\|_{1}$, as derived in the following lemma:
\begin{lemma} \label{lem:ae}
    \begin{equation} \label{eq:ae}
     \|\boldsymbol{\R}\|_{1} \leq \sum_{k=1,k\neq j}^{m}\|s_k\|_{1} \cdot \|\N(\mathbf{Z}_{k}^{(\ell)})\|_{1} + c,
     \end{equation}
     where $c \geq 0$ is a constant with respect to $\mathbf{s}$.
\end{lemma}
\begin{proof}
See Appendix.
\end{proof}}
Furthermore, since our intention is to participate as many preserved filters as possible in restoring a pruned filter, it would be better for the $s_k$ values to be more uniformly distributed. Both purposes can be fulfilled by this L2-based regularization term. 

\begin{algorithm}[t]
  \small
  \caption{LBYL method}
  \label{alg:Recovery}
  \KwIn{$\mathbf{\Theta}$ = $[\mathbf{W}^{{(1)}}$, .... ,$\mathbf{W}^{{(L)}}] := $ Original model}
  \KwOut{$\hat{\mathbf{\Theta}}$ = $[{\mathbf{\hat{W}}}^{{(1)}}$, .... ,$\mathbf{\hat{W}}^{(L)}] : = $ Pruned \& restored model}
   
  \For{each layer $\ell \in [1, L]$ that has been pruned}%$\in$ $\mathbb{R}^{n_{l} \times k \times k}$   $\mathbf{in}$ $\mathbf{W}^{(\ell)}$}
    {
    ${\boldsymbol{\S}}^{(\ell)} \leftarrow $ a zero-valued matrix
    
    \For{each filter $\mathbf{W}_i^{(\ell)}$ s.t. $i$ $\in$ $[1,m]$}{
        \If{$\mathbf{W}_i^{(\ell)}$ is not pruned} 
            {
            $\boldsymbol{\S}_{i,k}^{(\ell)} = 
              \begin{cases} 
              1  & \text{if  } i = i'_k \\
              0  & \text{otherwise } \\
              \end{cases}
             $
            }
        \Else{        
        $\boldsymbol{\S}_{i,:}^{(\ell)} \leftarrow {\argmin\limits_{\mathbf{s}}~\|\boldsymbol{\E}\|_2^{2} + \lambda_1 \|\boldsymbol{\B}\|_2^{2} + \lambda_2 \|\mathbf{s}\|_{2}^{2}}$  \tcp{obtained by Theorem \ref{thm:closedform}} 
        }
      }
    $\hat{\mathbf{W}}^{(\ell+1)} = \mathbf{W}^{(\ell+1)} \times_2 \boldsymbol{\S}^{(\ell)^T} $
    }
\Return $\hat{\mathbf{\Theta}}$
\end{algorithm}

\smalltitle{One-to-one vs. LBYL}
In one-to-one compensation approaches \cite{NM,Data-free}, they deliver the missing information for a pruned filter only to the most similar one of the remaining filters as shown in Figure \ref{fig:matrix:c}. Thus, this assumption is identical to the case that we have only one $s_{k}$ in Eq. (\ref{eq:goal2}), which would return a quite rough solution for the problem of minimizing the reconstruction error. Indeed, this one-to-one approach particularly fails to reduce the RE term as shown in Figure \ref{fig:error_components} of our experiments.

\begin{figure}[t!]
   \centering 
    \includegraphics[width=0.6\columnwidth]{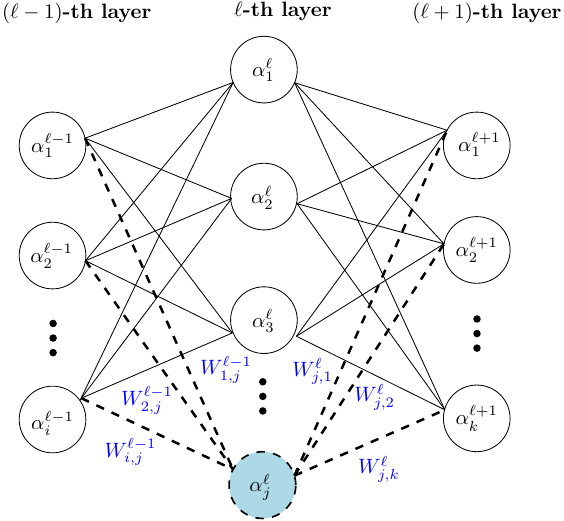}
   \caption{A neuron pruning scenario in fully-connected layers}
   \label{fig:appendix_figure}
\end{figure}

\begin{table*}[t] % 
\centering
{\small 
\begin{tabular}{c||c|c|c||c|c|c||c|c|c||c|c|c}  \Xhline{2\arrayrulewidth}
\multicolumn{13}{c}{\textbf{VGG-16 (Acc. 93.7)}}
\\ \Xhline{2\arrayrulewidth} %\hline
Criterion & \multicolumn{3}{c||}{L2-norm} & \multicolumn{3}{c||}{L2-GM} & \multicolumn{3}{c||}{L1-norm}& \multicolumn{3}{c}{Random}\\ \hline
Pruning Ratio& Ours& NM& Prune& Ours& NM& Prune& Ours& NM& Prune& Ours& NM& Prune\\ \Xhline{2\arrayrulewidth}
10\%& \textbf{92.04} & 91.93 & 89.43 & \textbf{92.15} & 91.82 & 88.83 & 91.85 & \textbf{91.97} & 89.85 & \textbf{88.05} & 86.52 & 74.87 \\\hline
20\%& \textbf{87.84} & 87.24 & 71.77 & \textbf{87.97} & 86.42 & 72.04 & \textbf{89.51} & 86.93 & 73.64 & \textbf{75.24} & 63.33 & 37.19 \\ \hline
30\%& \textbf{83.25} & 76.91 & 56.95 & \textbf{82.83} & 75.58 & 52.40 & \textbf{81.80} & 75.01 & 49.79 & 45.67 & \textbf{46.86} & 12.81 \\ \hline
40\%& \textbf{66.81} & 54.32 & 31.74 & \textbf{67.19} & 52.24 & 30.46 & \textbf{56.63} & 52.54 & 17.78 & \textbf{31.34} & 26.79 & 10.00 \\ \hline
50\%& \textbf{45.71} & 32.58 & 12.37 & \textbf{49.46} & 32.13 & 11.71 & \textbf{43.32} & 27.87 & 10.43 & \textbf{21.56} & 18.08 & 10.00 \\ \Xhline{2\arrayrulewidth}
% Average Acc. & \textbf{75.13} & 68.60 & 52.45 & \textbf{75.92} & 67.64 & 51.09 & \textbf{72.62} & 66.86 & 48.30 & \textbf{52.37} & 48.32 & 28.97\\ \Xhline{2\arrayrulewidth}   
\end{tabular}
}
\vspace{2mm}
\caption{Recovery results of VGG-16 on CIFAR-10}
\label{tab:VGG16-CIFAR10}
\vspace{1.5mm}

\centering 
{\small
\begin{tabular}{c||c|c|c||c|c|c||c|c|c||c|c|c}  \Xhline{2\arrayrulewidth}
\multicolumn{13}{c}{\textbf{ResNet-50 (Acc. 78.82) }}
\\ \Xhline{2\arrayrulewidth} %\hline
Criterion & \multicolumn{3}{c||}{L2-norm} & \multicolumn{3}{c||}{L2-GM} & \multicolumn{3}{c||}{L1-norm}& \multicolumn{3}{c}{Random}\\ \hline
Pruning Ratio& Ours& NM& prune& Ours& NM& prune& Ours& NM& prune& Ours& NM& prune
\\ \Xhline{2\arrayrulewidth}
10\%& \textbf{78.14} & 77.28 & 75.14 & \textbf{78.01} & 76.92 & 74.49 & \textbf{78.25} & 77.21 & 75.07 & \textbf{76.53} & 72.46 & 59.32 \\ \hline
20\%& \textbf{76.15} & 72.73 & 63.39 & \textbf{76.36} & 72.32 & 63.54 & \textbf{75.71} & 72.24 & 61.84 & \textbf{73.29} & 59.44 & 19.27 \\ \hline
30\%& \textbf{73.29} & 64.47 & 39.96 & \textbf{72.97} & 64.01 & 39.01 & \textbf{72.07} & 63.07 & 35.77 & \textbf{69.05} & 40.42 & 3.25  \\ \hline
40\%& \textbf{65.21} & 46.40 & 15.32 & \textbf{65.78} & 46.17 & 13.14 & \textbf{62.64} & 45.98 & 12.59 & \textbf{59.49} & 20.89 & 2.59  \\ \hline
50\%& \textbf{52.61} & 25.98 & 5.22  & \textbf{54.04} & 23.44 & 4.32  & \textbf{49.07} & 21.98 & 4.25  & \textbf{43.51} & 8.77  & 2.77  \\ \Xhline{2\arrayrulewidth}
Average Acc. & \textbf{69.08} & 57.37 & 39.81 & \textbf{69.43} & 56.57 & 38.90 & \textbf{67.55} & 56.10 & 37.90 & \textbf{64.37} & 40.40 & 17.44\\ \Xhline{2\arrayrulewidth}   
\end{tabular}%
}
\vspace{2mm}
\caption{Recovery results of ResNet-50 on CIFAR-100}
\label{tab:ResNet50-CIFAR100}

\end{table*}

\subsection{Our Train-Free Recovery Method with a Closed Form Solution}
Based on our data-free loss function in Eq. (\ref{eq:loss}), we now present our proposed train-free recovery method, called LBYL. First of all, we theoretically prove that the problem of minimizing $\mathcal{L}_{re}$ of Eq. (\ref{eq:loss}) has a closed form solution at the following theorem.
\begin{theorem} \label{thm:closedform}
The loss function $\mathcal{L}_{re}$ of Eq. (\ref{eq:loss}) is convex with a unique optimal solution as follows:
\begin{equation}
% \small
    \boldsymbol{s}=[X^{T}X+\lambda_{1}\frac{\gamma_{j}^2}{\sigma_{j}^2} \boldsymbol{p}\boldsymbol{p}^{T}+\lambda_{2}I]^{-1}[X^{T}\boldsymbol{y}+\frac{\lambda_{1}\gamma_{j}}{\sigma_{j}}(\frac{\mu_{j}\gamma_{j}}{\sigma_{j}}-\beta_{j})\boldsymbol{p}],
    \nonumber
\end{equation}
where (1) $X = [\frac{\sigma_{j}}{\gamma_{j}}\frac{\gamma_{1}}{\sigma_{1}}\boldsymbol{f_{1}}~~...~~\frac{\sigma_{j}}{\gamma_{j}}\frac{\gamma_{j-1}}{\sigma_{j-1}}\boldsymbol{f_{j-1}}~\frac{\sigma_{j}}{\gamma_{j}}\frac{\gamma_{j+1}}{\sigma_{j+1}}\boldsymbol{f_{j+1}}~~...~~ \\
\frac{\sigma_{j}}{\gamma_{j}}\frac{\gamma_{m}}{\sigma_{m}}\boldsymbol{f_{m}}]$ such that $\boldsymbol{f_{i}}$ is the vectorized $i$-th filter in the $\ell$-th layer, (2) $\boldsymbol{y}$ is the vectorized $j$-th filter in the $\ell$-th layer, and (3) $\boldsymbol{p} =  [p_{1}~...~p_{j-1}~p_{j+1}~...~p_{m}]^T$ such that $p_{i} ={\frac{\sigma_{j}}{\gamma_{j}}\frac{\gamma_{i}}{\sigma_{i}}}{(\mu_{i}-\frac{\sigma_{i}}{\gamma_{i}}\beta_{i})}$.
% where (1) $X = [~...~\frac{\sigma_{j}}{\gamma_{j}}\frac{\gamma_{j-1}}{\sigma_{j-1}}\boldsymbol{f_{j-1}}~\frac{\sigma_{j}}{\gamma_{j}}\frac{\gamma_{j+1}}{\sigma_{j+1}}\boldsymbol{f_{j+1}}~...~\frac{\sigma_{j}}{\gamma_{j}}\frac{\gamma_{m}}{\sigma_{m}}\boldsymbol{f_{m}}]$ such that $\boldsymbol{f_{i}}$ is the vectorized $i$-th filter in the $\ell$-th layer, (2) $\boldsymbol{y}$ is the vectorized $j$-th filter in the $\ell$-th layer, and (3) $\boldsymbol{p} =  [p_{1}~...~p_{j-1}~p_{j+1}~...~p_{m}]^T$ such that $p_{i} ={\frac{\sigma_{j}}{\gamma_{j}}\frac{\gamma_{i}}{\sigma_{i}}}{(\mu_{i}-\frac{\sigma_{i}}{\gamma_{i}}\beta_{i})}$.
\end{theorem}
\begin{proof}
See Appendix.
\end{proof}

Finally, given $\lambda_{1}$ and $\lambda_{2}$, Algorithm \ref{alg:Recovery} presents the whole procedure of our LBYL method using the closed form solution in Theorem \ref{thm:closedform}.

\smalltitle{Extension of LBYL to neuron pruning}
Even though this article focuses on filter pruning, our LBYL method can also be applied to prune and restore neurons in vanilla feed-forward neural networks consisting of only FC layers such as LeNet-300-100 \cite{LeNet}. Let $\alpha_{i}^{\ell}$ denote the output of the $i$-th neuron in the $\ell$-th layer, and let $W^{\ell}_{j,k}$ denote a weight between the $j$-th neuron in the $\ell$-th layer and the $k$-th neuron in the $(\ell+1)$-th layer as illustrated in Figure \ref{fig:appendix_figure}. The goal of our method is still minimizing the reconstruction error between a restored network and its original network by making a linear combination of remained neurons to compensate each pruned neuron. If the $j$-th neuron in the $\ell$-th layer is pruned, the reconstruction error can be formulated as: 
\begin{equation}
\sum\limits_{n = 1}^{k}\|{{\mathbf{\alpha}}_{n}^{{\ell+1}}-{\hat{\mathbf{\alpha}}}_{n}^{{\ell+1}}}\|_1 
\nonumber
\end{equation}
As explained in Section \ref{sec:probdef}, the reconstruction error for the $k$-th neuron at the $(\ell+1)$-th layer can be derived as:
\begin{eqnarray}
 \|{\mathbf{\alpha}}_{k}^{{\ell+1}} - (\sum\limits_{n = 1, n \neq j}^{j}{ W^{\ell}_{n,k} * \mathbf{\alpha}_{n}^{\ell}} +{\sum\limits_{n = 1, n \neq j}^{j} s_nW^{\ell}_{j,k} *  \mathbf{\alpha}_{n}^{\ell}} ) \|_1 
 \\ 
 = \|{ W_{j,k}^{\ell} * (\mathbf{\alpha}_{j}^{(\ell)} - \sum\limits_{n = 1, n \neq j}^{j}{s_{n}} \mathbf{\alpha}_{n}^{(\ell)})}\|_1 ~~~~~~~~~~~
% = \|{ W^{\ell}_{j,n} * (\sum\limits_{n' = 1, n' \neq j}^{j}{s_{n'}}\mathbf{\alpha}_{n'}^{\ell})}\|_1 ,
\nonumber
\end{eqnarray}
where we need to find scalars $s_{n}$'s such that ${ \mathbf{\alpha}_{j}^{(\ell)} - \sum\limits_{n = 1, n \neq j}^{j}{s_{n}} \mathbf{\alpha}_{n}^{(\ell)}}$ to minimize the reconstruction error. Note that $\alpha$ is data dependent and computed through a multiplication procedure with weights followed by an activation function (e.g., ReLU) without batch normalization process in LeNet-300-100. Therefore, by the similar derivation of Section \ref{sec:method:a}, we have:
\begin{equation}
     \mathcal{L}_{re} =  \|\boldsymbol{\E}\|_2^{2} + \lambda \|\mathbf{s}\|_{2}^{2}, ~~\text{where}~\mathbf{s} =  [s_{1}~...~s_{j-1}].
    \nonumber
\end{equation}

%% file: experiments.tex
\section{Experiments} \label{sec:experiments}
In this section, we empirically validate the performance and effectiveness of our proposed LBYL method, compared to a one-to-one compensation method, called \textit{Neuron Merging} (\textit{NM})  \cite{NM} as well as a baseline called \textit{Prune} that does not perform any recovery process after pruning. For a fair comparison, we exploit the code provided by the authors of \cite{NM} and then implement our proposed method using PyTorch \cite{PyTorch}. LBYL uses two hyperparameters, namely $\lambda_1$ and $\lambda_2$, that adjust the weights of loss terms BE and AE, respectively. The details of hyperparameter settings are presented in Appendix. Also, our implementation is available at our github.\footnote{https://github.com/bigdata-inha/LBYL}

\smalltitle{Pruning criteria and ratio}
\rev{We evaluate LBYL using the following four data-independent pruning criteria: L1-norm \cite{PFEC}, L2-norm \cite{Soft}, L2-GM \cite{FPGM}, and lastly even random. Random pruning is intended to show the robustness of each method when a norm-based criterion is not used. We prune only convolution filters in CNNs, and hence the pruning ratio is the ratio of pruned filters for each convolution layer.}

\subsection{Experiments Using CIFAR-10 and CIFAR-100}
\smalltitle{Dataset and pretrained model} 
The CIFAR-10 dataset \cite{krizhevsky2009learning} consists 60K images of 10 different classes, where each class includes 5K training images along with 1K validation images and the size of every image is 32 × 32. CIFAR-100 \cite{krizhevsky2009learning} also include 60K images for 100 different classes, where 500 training images along with 100 validation images are included for each class. We test VGG-16 \cite{VGG} for CIFAR-10 and ResNet-50 \cite{ResNet} for CIFAR-100, and present the result of CIFAR-100 only in Appendix as 
their experimental results of CIFAR-10 and CIFAR-100 show the similar trend.

We adopt a pretrained VGG-16 released by NM \cite{NM} in the experiments on CIFAR-10. For CIFAR-100, we use ResNet-50 \cite{ResNet} that we train from scratch in a way that:
(1) SGD with Nesterov momentum of 0.9 is used, (2) the initial learning rate is 0.1 and then divided by 5 at 60, 120, and 160 epochs, (3) the weight decay factor is set to 5e-4, and (4) the total duration of training is 200 epochs with the batch size 128.

\begin{table*}[htb!]
% \begin{table}[t]
{\small

\begin{tabular}{c||c|c|c||c|c|c||c|c|c||c|c|c}  \Xhline{2\arrayrulewidth}
\multicolumn{13}{c}{\textbf{ResNet34 - ImageNet (Acc. 73.27)}}
\\ \Xhline{2\arrayrulewidth} %\hline
Criterion & \multicolumn{3}{c||}{L2-norm} & \multicolumn{3}{c||}{L2-GM} & \multicolumn{3}{c||}{L1-norm}& \multicolumn{3}{c}{Random}\\ \hline
Pruning Ratio& Ours& NM& Prune& Ours& NM& Prune& Ours& NM& Prune& Ours& NM& Prune\\ \Xhline{2\arrayrulewidth}
10\%& \textbf{69.22} & 66.96 & 63.74 & \textbf{69.12} & 66.59 & 61.76 & \textbf{69.07} & 66.30 & 62.05 & \textbf{65.90} & 64.75& 52.97 \\ \hline
20\%& \textbf{62.49} & 55.70 & 42.81 & \textbf{62.27} & 55.39 & 43.45 & \textbf{61.18} & 53.95 & 40.61 & \textbf{50.64} & 48.40& 18.62 \\ \hline
30\%& \textbf{47.59} & 39.22 & 17.02 & \textbf{49.43} & 37.97 & 15.74 & \textbf{46.88} & 35.56 & 12.58 & 22.92 & \textbf{23.69} & 1.35  \\ \Xhline{2\arrayrulewidth}
% Average Acc. & \textbf{59.77} & 53.96 & 41.19 & \textbf{60.27} & 53.32 & 40.32 & \textbf{59.04} & 51.94 & 38.41 & \textbf{46.49} & 45.61& 19.10\\ \Xhline{2\arrayrulewidth}
\end{tabular}
}
\vspace{2mm}
\caption{Recovery results of ResNet-34 on ImageNet}
\label{tab:ResNet34-ImageNet}
\vspace{1.5mm}

% \end{table}
% \begin{table}[t]
{\small

\begin{tabular}{c||c|c|c||c|c|c||c|c|c||c|c|c}  \Xhline{2\arrayrulewidth}
\multicolumn{13}{c}{\textbf{ResNet101 - ImageNet (Acc. 77.31)}}
\\ \Xhline{2\arrayrulewidth} %\hline
Criterion & \multicolumn{3}{c||}{L2-norm} & \multicolumn{3}{c||}{L2-GM} & \multicolumn{3}{c||}{L1-norm}& \multicolumn{3}{c}{Random}\\ \hline
Pruning Ratio& Ours& NM& Prune& Ours& NM& Prune& Ours& NM& Prune& Ours& NM& Prune
\\ \Xhline{2\arrayrulewidth}
10\%& \textbf{74.59} & 72.36 & 68.90 & \textbf{74.54} & 72.38 & 68.38 & \textbf{74.19} & 72.05 & 68.33 & \textbf{59.78} & 57.13 & 46.55 \\ \hline
20\%& \textbf{68.47} & 61.42 & 45.78 & \textbf{69.03} & 61.02 & 46.25 & \textbf{68.52} & 60.57 & 44.57 & \textbf{34.85} & 15.78 & 6.83  \\ \hline
30\%& \textbf{55.51} & 37.38 & 10.32 & \textbf{56.65} & 33.74 & 7.65  & \textbf{56.11} & 36.43 & 9.32  & \textbf{13.27} & 2.31  & 0.60  \\ \Xhline{2\arrayrulewidth}
% Average Acc. & \textbf{66.19} & 57.05 & 41.67 & \textbf{66.74} & 55.71 & 40.76 & \textbf{66.27} & 56.35 & 40.74 & \textbf{35.97} & 25.07 & 17.99\\\Xhline{2\arrayrulewidth}
\end{tabular}
}
\vspace{2mm}
\caption{Recovery results of ResNet-101 on ImageNet} \label{tab:ResNet101-ImageNet}
\vspace{1.5mm}
% \end{table}

% \begin{table}[h]
\centering
\small
\begin{tabular}{c||c |c |c ||c |c |c ||c |c |c } \Xhline{2\arrayrulewidth}
\multicolumn{10}{c}{\textbf{MobileNet-V2 (Acc. 71.88) }} \\ \Xhline{2\arrayrulewidth} %\hline
\multicolumn{1}{c||}{Criterion} & \multicolumn{3}{c||}{ L2 - norm} & \multicolumn{3}{c||}{ L2 - GM} & \multicolumn{3}{c}{L1 - norm} \\ \hline
Pruning Ratio & Ours & NM & prune & Ours & NM & prune & Ours & NM & prune \\ \Xhline{2\arrayrulewidth}
5\% & \textbf{67.46} & 67.1 & 61.97 & \textbf{67.64} & 63.68 & 62.53 & 67.25 & \textbf{67.64} & 63.27 \\ \hline
10\% & \textbf{53.41} & 52.66& 29.06 & \textbf{52.26} & 46.8& 28.21 & \textbf{54.95} & 52.1& 32.08 \\ \Xhline{2\arrayrulewidth}
\end{tabular}%
\vspace{2mm}
\caption{Recovery results of MobileNet-V2 on ImageNet}
\label{tab:mobilenet}
\end{table*}

% In case of CIFAR-100, we evaluate the our proposed method using the ResNet-50 \cite{ResNet}. To train the ResNet-50, we exploit SGD with Nesterov momentum of 0.9. the initial learning rate is 0.1 and it is divided by 5 at 60, 120, 160 epochs. the weight decay is set to 5e-4, the train epochs is 200 with batch size 128. 

\smalltitle{Results}
We first prune VGG-16 on CIFAR-10, which is a representative single-branched CNN, using the layer-by-layer pruning scheme with four different criteria, and then update the remaining filters in each layer to restore the pruned VGG-16. As shown in Table \ref{tab:VGG16-CIFAR10}, our LBYL method achieves up to 6.16\% higher average accuracy than NM \cite{NM}. In particular, LBYL outperforms NM with clear margins when the pruning ratio increases. This is probably because NM might have trouble finding a filter sufficiently close to each pruned filter with less remaining filters due to a higher pruning ratio. Table \ref{tab:ResNet50-CIFAR100} presents the experimental results of CIFAR-100 on ResNet-50. In all the cases, it is surely confirmed that our LBYL outperforms NM with clear margins.

\subsection{Experiments Using ImageNet (ILSVRC2012)}

\smalltitle{Dataset and pre-trained model} 
The ImageNet dataset \cite{deng2009imagenet} contains 1,000 classes and consist of 1.28M training images and 50K validation images of size 256 $\times$ 256. For ImageNet, we use pretrained ResNet-34 and ResNet-101, both of which are released by PyTorch\footnote{https://pytorch.org/vision/stable/models.html}. ResNet-34 and ResNet-101 respectively represent two different types of the ResNet architecture, namely the one consisting of basic blocks and the other one consisting of bottleneck blocks.

\smalltitle{Results}
To prune a ResNet architecture, we exploit the common strategy to remove each residual block, that is, pruning the first two convolutional layers and not changing the output dimension of the block. Tables \ref{tab:ResNet34-ImageNet} and \ref{tab:ResNet101-ImageNet} present the experimental results using ResNet-34 and ResNet-101 on ImageNet. Through the results, LBYL can still recover the damaged filters to a satisfactory extent.

\smalltitle{Result of MobileNet-V2 on ImageNet}
Table \ref{tab:mobilenet} shows the experimental results of ImageNet on MobileNet-V2, where we prune only the first layer of each block. This scheme can be seen as a naive adaptation of our method for MobileNet-V2, but LBYL still manages to beat NM in such a tiny architecture.

\begin{figure*}[t]
	\centering 
	\includegraphics[height=2mm]{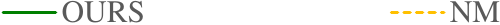} \\
    \subfigure[\label{fig:expcos_:a}{RE} ]{\hspace{0mm}\includegraphics[width=0.45\columnwidth]{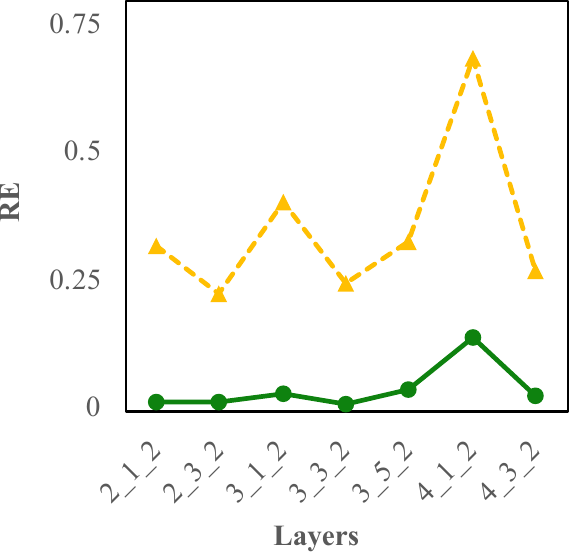}\hspace{2mm}}
    \subfigure[\label{fig:expcos_:b}{BE} ]{\hspace{0mm}\includegraphics[width=0.45\columnwidth]{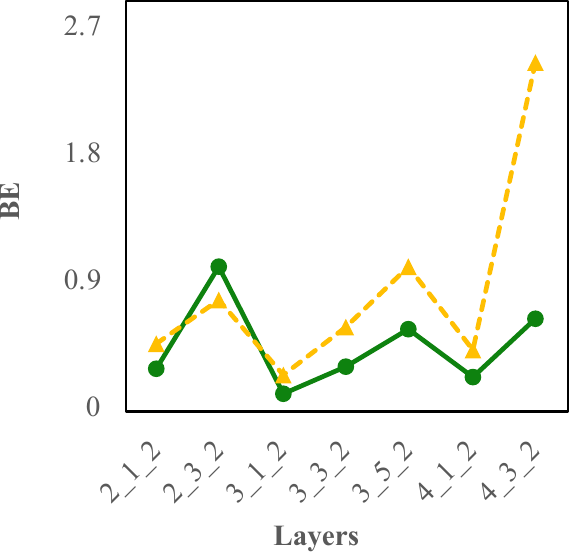}\hspace{2mm}}
    \subfigure[\label{fig:expcos_:d}{WARE} ]{\hspace{0mm}\includegraphics[width=0.45\columnwidth]{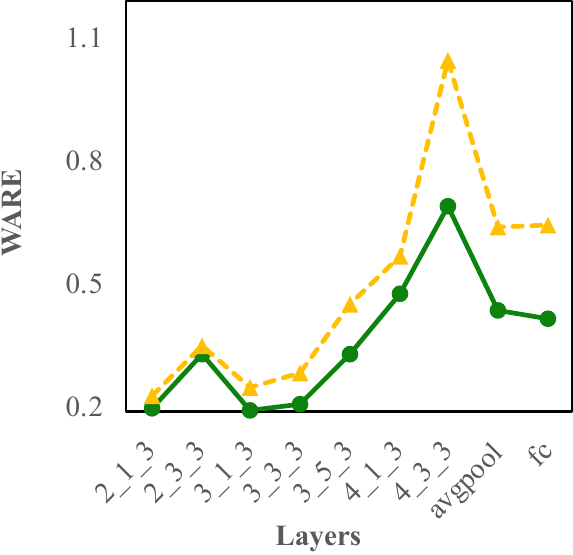}\hspace{2mm}}
    \subfigure[\label{fig:expcos_:c}{Average Norm of Scales} ]{\hspace{0mm}\includegraphics[width=0.45\columnwidth]{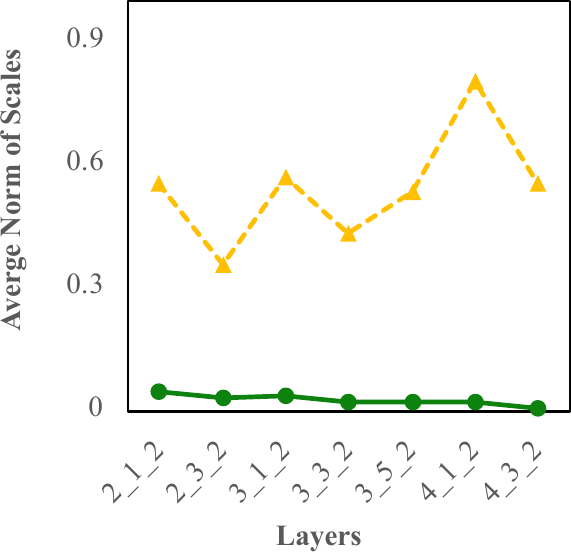}\hspace{2mm}}
    \caption{Comparison on the three error components with NM \cite{NM}, where each $m$\_$n\_k$ in the x-axis represents the $k$-th conv module in the $n$-th block at the $m$-th layer in ResNet-50 and when pruning the first and second convolution layers of each block by 30\%}
	\label{fig:error_components}
	\vspace{-2mm}
\end{figure*}

\subsection{\revfour{Experiments Using COCO2017}}

\revfour{

\smalltitle{Dataset and pre-trained model}
The COCO2017 dataset \cite{cocodataset} is widely used for training and evaluating object detection, segmentation, and captioning tasks, which contains 330K training images alongside 5K validation images across 80 object categories. We use the SSD (Single Shot Multibox Detector \cite{ssd}) model based on ResNet-50 for evaluating the object detection task on COCO2017, obtained from a PyTorch implementation library \footnote{https://github.com/NVIDIA/DeepLearningExamples/tree/master/PyTorch}. 

\smalltitle{Results}
In the SSD detector, we prune the ResNet-50 backbone using the same pruning criterion as in the experiments using ImageNet, and measure the recovered performance of pruned ResNet-50 by each compared recovery method. In terms of implementation, here we do not utilize the Batch Normalization Error (BE) as the values of $\gamma$ and $\sigma$ of batch normalization layers of the pre-trained SSD model are too tiny (less than 10$^{-6}$), which incurs zero division error in the implementation level. Table \ref{tab:COCO_res} demonstrates the experimental results on the recovered performance of the pruned SSD model on COCO2017. Our LBYL method shows better performance than NM \cite{NM} by clear margins. More specifically, as the pruning ratio increases, LBYL achieves up to approximately 5\% higher AP$_{50}$ on average than NM.
}

% Also, we set $\lambda$ to 0.05 in the recovery process to show the robustness of lambda value in various pruning criteria.

% \begin{table}[t!]
% \color{blue}
% \centering
% \small
% \begin{tabular}{c||c|c|c} \Xhline{2\arrayrulewidth}
% \multicolumn{4}{c}{\textbf{SSD based on ResNet-50 on COCO2017 (AP 25.35)}} \\
% \Xhline{2\arrayrulewidth}
% Pruning Ratio& Ours & NM& Prune\\ \Xhline{2\arrayrulewidth}
% 10 \% & \textbf{22.62}   & 22.19  &20.70 \\\hline
% 20 \% & \textbf{17.10}  & 15.47  &12.69 \\\hline 
% 30 \% & \textbf{8.22}  & 5.95  &4.33 \\\Xhline{2\arrayrulewidth}
% \end{tabular}
% \vspace{3mm}
% \caption{ Average Precisions of Single Shot Object Detetion based on ResNet-50 on COCO2017 using l2-norm pruning criterion. }
% \label{tab:COCO}
% \end{table}

\begin{table*}[htb!]
\centering
\small
% \color{blue}
\begin{tabular}{cc||c|c|c|c|c|c|c}
\hline \Xhline{2\arrayrulewidth}

\multicolumn{2}{c||}{} & Method & AP& AP$_{50}$ & AP$_{75}$& AP$_{S}$& AP$_{M}$ & AP$_{L}$          \\ \hline  \Xhline{2\arrayrulewidth}
\multicolumn{1}{l|}{Pruning   Criterion}      & Pruning Ratio         & SSD (ResNet-50)    & 25.35              & 42.75            & 26.19              &       7.13        & 27.31              & 40.88             \\ \hline \Xhline{2\arrayrulewidth}
\multicolumn{1}{c|}{\multirow{9}{*}{L2-norm}} & \multirow{3}{*}{10\%} & Prune  & 20.7           & 37.08          & 20.64          & 6.38          & 22.3           & 32.46          \\ \cline{3-9} 
\multicolumn{1}{c|}{}                         &                       & NM     & 22.2           & 38.27          & 22.53          & 6.31          & 23.58          & 36.02          \\ \cline{3-9} 
\multicolumn{1}{c|}{}                         &                       & Ours   & \textbf{22.63} & \textbf{39.83} & \textbf{22.85} & \textbf{6.5}  & \textbf{24.02} & \textbf{36.37} \\ \cline{2-9} 
\multicolumn{1}{c|}{}                         & \multirow{3}{*}{20\%} & Prune  & 12.69          & 24.38          & 12.06          & 3.8           & 15.14          & 19.51          \\ \cline{3-9} 
\multicolumn{1}{c|}{}                         &                       & NM     & 15.47          & 27.39          & 15.37          & 3.72          & 16.26          & 25.53          \\ \cline{3-9} 
\multicolumn{1}{c|}{}                         &                       & Ours   & \textbf{17.11} & \textbf{32.23} & \textbf{16.39} & \textbf{5.05} & \textbf{18.36} & \textbf{27.66} \\ \cline{2-9} 
\multicolumn{1}{c|}{}                         & \multirow{3}{*}{30\%} & Prune  & 4.33           & 9.32           & 3.58           & 1.52          & 6.49           & 5.47           \\ \cline{3-9} 
\multicolumn{1}{c|}{}                         &                       & NM     & 5.95           & 11.51          & 5.59           & 1.52          & 7.34           & 8.73           \\ \cline{3-9} 
\multicolumn{1}{c|}{}                         &                       & Ours   & \textbf{8.22}  & \textbf{18.62} & \textbf{6.15}  & \textbf{2.59} & \textbf{9.86}  & \textbf{12.8}  \\   \hline \Xhline{2\arrayrulewidth}
\multicolumn{1}{c|}{\multirow{9}{*}{L1-norm}} & \multirow{3}{*}{10\%} & Prune  & 20.69          & 37.31          & 20.68          & 6.15          & 22.77          & 31.96          \\ \cline{3-9} 
\multicolumn{1}{c|}{}                         &                       & NM     & 22.1           & 38.16          & 22.54          & 5.98          & 23.27          & 35.71          \\ \cline{3-9} 
\multicolumn{1}{c|}{}                         &                       & Ours   & \textbf{22.63} & \textbf{39.66} & \textbf{22.97} & \textbf{6.45} & \textbf{24.11} & \textbf{36.15} \\ \cline{2-9} 
\multicolumn{1}{c|}{}                         & \multirow{3}{*}{20\%} & Prune  & 10.87          & 21.95          & 9.78           & 3.51          & 13.62          & 15.82          \\ \cline{3-9} 
\multicolumn{1}{c|}{}                         &                       & NM     & 14.59          & 25.6           & 14.39          & 3.5           & 15.6           & 23.59          \\ \cline{3-9} 
\multicolumn{1}{c|}{}                         &                       & Ours   & \textbf{16.06} & \textbf{31.18} & \textbf{15.08} & \textbf{4.95} & \textbf{17.28} & \textbf{25.77} \\ \cline{2-9} 
\multicolumn{1}{c|}{}                         & \multirow{3}{*}{30\%} & Prune  & 2.98           & 6.86           & 2.12           & 1.26          & 4.56           & 3.63           \\ \cline{3-9} 
\multicolumn{1}{c|}{}                         &                       & NM     & 5.85           & 11.57          & 5.36           & 1.59          & 7.07           & 7.97           \\ \cline{3-9} 
\multicolumn{1}{c|}{}                         &                       & Ours   & \textbf{7.1}   & \textbf{16.73} & \textbf{4.97}  & \textbf{2.39} & \textbf{9.07}  & \textbf{10.85} \\  \hline \Xhline{2\arrayrulewidth}
\multicolumn{1}{c|}{\multirow{9}{*}{L2-GM}}   & \multirow{3}{*}{10\%} & Prune  & 20.68          & 37.09          & 20.68          & 6.42          & 22.3           & 32.1           \\ \cline{3-9} 
\multicolumn{1}{c|}{}                         &                       & NM     & 22.11          & 38.06          & 22.53          & 6.18          & 23.42          & 36.03          \\ \cline{3-9} 
\multicolumn{1}{c|}{}                         &                       & Ours   & \textbf{22.88} & \textbf{39.96} & \textbf{23.35} & \textbf{5.59} & \textbf{24.5}  & \textbf{36.45} \\ \cline{2-9} 
\multicolumn{1}{c|}{}                         & \multirow{3}{*}{20\%} & Prune  & 12.98          & 24.78          & 12.16          & 3.99          & 15.2           & 20.16          \\ \cline{3-9} 
\multicolumn{1}{c|}{}                         &                       & NM     & 15.58          & 27.55          & 15.73          & 3.96          & 16.57          & 25.37          \\ \cline{3-9} 
\multicolumn{1}{c|}{}                         &                       & Ours   & \textbf{17.87} & \textbf{33.24} & \textbf{17.34} & \textbf{5.31} & \textbf{19.26} & \textbf{28.85} \\ \cline{2-9} 
\multicolumn{1}{c|}{}                         & \multirow{3}{*}{30\%} & Prune  & 4.17           & 9.15           & 3.3            & 1.51          & 6.21           & 5.35           \\ \cline{3-9} 
\multicolumn{1}{c|}{}                         &                       & NM     & 6.14           & 11.88          & 5.85           & 1.49          & 7.66           & 8.96           \\ \cline{3-9} 
\multicolumn{1}{c|}{}                         &                       & Ours   & \textbf{10.36} & \textbf{21.75} & \textbf{8.59}  & \textbf{3.17} & \textbf{11.65} & \textbf{16.3}  \\  \hline \Xhline{2\arrayrulewidth}
\end{tabular}
\vspace{3mm}
\caption{ Recovery Results of Single Shot Object Detetion based on ResNet-50 on COCO2017}
\label{tab:COCO_res}
\end{table*}

\subsection{Effectiveness in terms of Reconstruction}
In order to see the effectiveness of our loss function on minimizing the reconstruction error, we compute the values of three error components, namely RE, BE, and AE, using ResNet-50 \cite{ResNet} on CIFAR-100 \cite{krizhevsky2009learning}. Both RE and BE can be measured without any inference with sample data, but AE is not easy to obtain. Therefore, we adopt the \textit{weighted average reconstruction error} (\textit{WARE}) introduced in \cite{NISP}, which shows the difference between the layer-wise output of the original model and that of the pruned model.

Figure \ref{fig:error_components} shows layer-wise values for each of RE, BE and WARE. It is clearly observed that NM returns a very rough approximation in terms of RE, compared to our LBYL method. This is due to the fact that NM uses only a single remaining filter to receive the missing information of a pruned filter, and consequently the final output can be quite far from the original.
In both BE and WARE, there is not much difference between LBYL and NM in early layers, but we can observe that the error values of NM become higher in the end. Thus, even though NM might be effective to reconstruct the original output in some layers, it shows unstable performance depending on the existence of similar filters and finally ends up with less accurate approximation. On the other hand, our LBYL method is less varied across layers and managed to keep the error values lower in the end for all three metrics.

\revsec{Furthermore, we measure the average scale coefficients for each layer. As shown in Figure \ref{fig:expcos_:c}, the average norm of $\mathbf{s}$ of LBYL becomes smaller as the number of filters increases in deeper layers due to our regularization term $\|\mathbf{s}\|_{2}^2$, while the scale norm of NM gets rather larger in those deeper layers. This is somewhat consistent to the comparison result of WARE in Figure \ref{fig:expcos_:d}, where the performance gap between ours and NM gets larger in deeper layers.}

\begin{table*}[t]
\centering
% \color{blue}
\begin{tabular}{c||c|c|c|c|c} \Xhline{2\arrayrulewidth}
Pruning Ratio& Ours & NM& Coreset&Prune & From Scratch (80 epochs) \\ \Xhline{2\arrayrulewidth}
10 \% & \textbf{78.14 → 78.3}   & 77.28 → 77.57  & 50.28 → 76.55& 75.14 → 77.67  & 48.22 (72.07)\\\hline
20 \% & \textbf{76.15 → 77.22}  & 72.73 → 75.62  & 23.9 → 74.39& 63.39 → 75.4   & 47.85 (72.01)\\\hline 
30 \% & \textbf{73.29 → 75.99}  & 64.47 → 73.03  & 17.81 → 71.88&  39.96 → 73.18  & 46.7 (72.24)\\\hline
40 \% & \textbf{65.21 → 74.13}  & 46.4 → 69.71   & 6.13 → 68.14&  15.32 → 70.12  & 46.53 (72.88)\\\hline
50 \% & \textbf{52.61 → 71.15}  & 25.98 → 65.03  & 1.28 → 62.88&  5.22 → 65.72  & 44 (69.33)\\\Xhline{2\arrayrulewidth}
\end{tabular}
\vspace{2mm}
\caption{Fine-tuned or trained accuracies of ResNet-50 on CIFAR-100 using L2-norm criterion, where we fine-tune each model for 20 epochs and train the same-sized small architecture for 20 epochs (80 epochs)}
\label{tab:finetune_scratch}
\end{table*}

\begin{figure*}[t]
	\centering 
	\includegraphics[height=2mm]{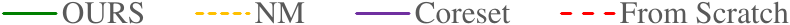} \\
    \subfigure[\label{fig:expcos:a}10\%]{\hspace{0mm}\includegraphics[width=0.45\columnwidth]{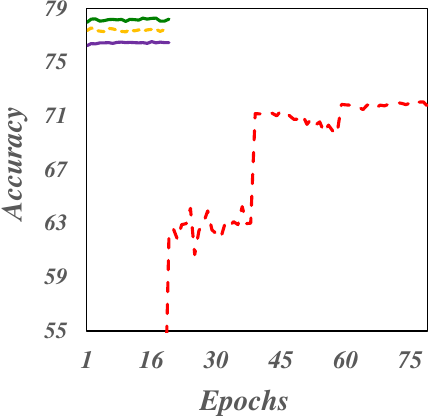}\hspace{2mm}}
    \subfigure[\label{fig:expcos:b}30\%]{\hspace{0mm}\includegraphics[width=0.45\columnwidth]{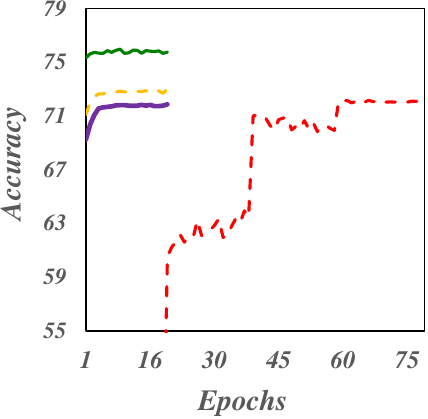}\hspace{2mm}}
    \subfigure[\label{fig:expcos:c}50\%]{\hspace{0mm}\includegraphics[width=0.45\columnwidth]{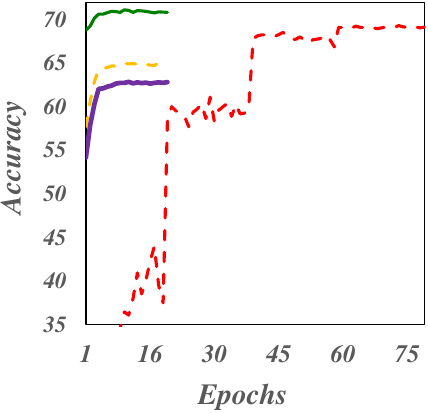}\hspace{2mm}}
    \vspace{2mm}
    \caption{Comparison on learning curves of fine-tuning restored networks for 20 epochs and that of training the same-sized small architecture from scratch for 80 epochs at different pruning ratios}
	\label{fig:Fine tuning and From Scratch}
\end{figure*}

\smalltitle{Performance estimation}
One can be curious about how to set the appropriate pruning ratio for a particular target size and accuracy if we cannot use any data during the entire pruning and recovery process. Our two data-free error terms (i.e., RE and BE) can also be utilized in such a case. As shown in Figure \ref{fig:error_components}, we can roughly estimate WARE, which is also highly relevant to the final performance, by combining RE and BE.

\subsection{Practical Test with Data and Fine-Tuning} \label{sec:praticaltest}
Although this article focuses on the problem of restoring pruned networks without data, we additionally test whether our method is also effective when the data becomes available later on. Table \ref{tab:finetune_scratch} shows the experimental results on the \revfour{initial and} final accuracy of either pruned or restored model after fine-tuning along with that of the same-sized architecture trained from scratch. In this experiment, we take the data-independent pruning method based on \textit{Coreset} \cite{CoreSet_ICLR} as another competitor, which also introduces a training-free recovery method as an intermediate process. \revfour{We examine how much the initially measured performance right after recovery has been improved with the help of light finetuning for 20 epochs, and also show the resulting accuracy of the model trained from scratch after 80 epochs.} In all cases, our LBYL method outperforms its competitors particularly when the pruning ratio increases. Our best guess on why NM becomes eventually worse than just pruned model is that NM can choose a dissimilar filter to recover each pruned one in some layers, which makes it hard for fine-tuning to improve the performance. Coreset \cite{CoreSet_ICLR} shows the lowest accuracy probably because it does not focus on training-free recovery and hence 20 epochs of fine-tuning is not sufficient. As shown in Figure \ref{fig:Fine tuning and From Scratch}, \revfour{LBYL shows the fastest convergence speed during the fine-tuning process to the point that it takes only a few epochs to reach the maximum accuracy. Both of these observations justify the fact that our LBYL method is not just theoretically rigorous but also practically meaningful when the data is available.}

\begin{table*}[t]
\centering
\small
\begin{tabular}{c||c|c|c}
\hline \Xhline{2\arrayrulewidth}
Method        & FLOPs (G) & Top-1 Acc (\%) &  Time for Pruning + Fine-Tuning (epochs)\\ \hline \Xhline{2\arrayrulewidth}
ResNet-50 (Original) & 4.1    & 76.1 & - \\ \hline \Xhline{2\arrayrulewidth}
DeepInversion\cite{DeepInversion} (w/ 0.1M ImageNet) & 2.9    & \textbf{74.9} & $\leq$ 22 (2.66 GFlops) + 30  \\ \hline
Ours (w/ 0.1M ImageNet)    &      2.9  &   70.9 & \textbf{0 + 1}     \\ \hline \Xhline{2\arrayrulewidth}
% DeepInversion\cite{DeepInversion} & 2.9    & 73.3 & 35673.6 + \underline{$\alpha$} \\ \hline
% Ours (w/o fine-tuning )       &    2.9    &   49.7 & 16.1   \\ \hline
% \Xhline{2\arrayrulewidth}
\end{tabular}
\vspace{2mm}
\caption{Fine-tuned accuracies of ResNet-50 on partial ImageNet}
\label{tab:finetune_synthetic}
\end{table*}

% \begin{table*}[t]
% \centering
% \small
% \begin{tabular}{c||c}
% \hline \Xhline{2\arrayrulewidth}
% Method        & latency (sec) \\ \hline \Xhline{2\arrayrulewidth}
% Red++\cite{Red++} & 5.01 \\ \hline
% OURS  &       2.08 (x2.41)  \\ \hline
% \Xhline{2\arrayrulewidth}
% \end{tabular}
% \vspace{2mm}
% \caption{Latency test on CPU (Intel Core Xeon Gold5122 ).}
% \label{tab:inference_res}
% \end{table*}

\subsection{Comparison with SOTA Data-Free Compression Methods}
\rev{
Since this article focuses on the problem of restoring pruned networks without any retraining, it is not trivial to make a fair comparison with data-free compression methods fully supported by retraining. Among various data-free methods, we compare our method with DeepInversion \cite{DeepInversion}, as it also utilizes pruning together with retraining process using a generator. DeepInversion fine-tunes a pruned network using synthetic data that is generated by its own generator, which however cannot be used to fine-tune our restored networks as the generator is tailored to the iterative pruning strategy of DeepInversion. Therefore, we employ a partial training dataset (0.1M ImageNet) to fine-tune either the network pruned by DeepInversion or our restored network. As shown in Table \ref{tab:finetune_synthetic}, our methods shows 4\% drop in accuracy, compared to DeepInversion. In fact, this performance gap comes from their different pruning schemes, namely one-shot pruning for ours and iterative pruning for DeepInversion. Thus, our LBYL method performs one-shot pruning for a pretrained model, and then we fine-tune the recovered model only for one epoch to achieve 70.9\% accuracy, while DeepInversion iteratively prunes filters during fine-tuning, which usually takes a long time to reach to a target compression ratio (e.g., 10 epochs for 3.27 GFlops and 22 epochs for 2.66 GFlops \cite{DeepInversion}).
}

\subsection{\revfour{Neuron Version of LBYL with LeNet-300-100}}
%  Even though this paper focuses on filter pruning, our LBYL method can also be applied to prune and restore neurons in vanilla feed-forward neural networks consisting of only FC layers. We present the details in Appendix due to the space limitation
\revfour{As discussed earlier, our LBYL method can also be extended to \textit{neuron pruning}, instead of filter pruning, in a vanilla feed-forward neural network such as LeNet-300-100 \cite{LeNet}. Table \ref{tab:LeNet_FashionMNIST} shows the experimental results of LeNet-300-100 on FashionMNIST by following the same setup of \textit{Coreset} \cite{CoreSet_ICLR}. It is well observed that LBYL outperforms the training-free recovery version of Coreset with clear margins. Although LBYL occasionally takes the second place, it still manages to achieve the best accuracy in overall.}

\begin{table*}[t!]
\centering
% \color{blue}
\begin{tabular}{c||ccc||ccc||ccc||c}
\Xhline{2\arrayrulewidth}
\multicolumn{11}{c}{\textbf{LeNet-300-100 (Acc. 89.51) }} \\ \Xhline{2\arrayrulewidth} %\hline
\hline Criterion& \multicolumn{3}{c||}{L2-norm}& \multicolumn{3}{c||}{L2-GM}& \multicolumn{3}{c||}{L1-norm}& \multirow{2}{*}{Coreset} \\ \cline{1-10}
Pruning Ratio& \multicolumn{1}{c|}{Ours}& \multicolumn{1}{c|}{NM}& Prune & \multicolumn{1}{c|}{Ours}& \multicolumn{1}{c|}{NM}& Prune & \multicolumn{1}{c|}{Ours}& \multicolumn{1}{c|}{NM}& Prune &\\ \hline \Xhline{2\arrayrulewidth}
50 \%& \multicolumn{1}{c|}{\textbf{88.83}} & \multicolumn{1}{c|}{87.86}          & 87.86 & \multicolumn{1}{c|}{\textbf{88.69}}  & \multicolumn{1}{c|}{88.57} & 88.08 & \multicolumn{1}{c|}{\textbf{89.03}} & \multicolumn{1}{c|}{88.69} & 88.40 & 79.34\\ \hline
60 \%& \multicolumn{1}{c|}{87.75}& \multicolumn{1}{c|}{\textbf{88.07}} & 83.03 & \multicolumn{1}{c|}{\textbf{88.15}}  & \multicolumn{1}{c|}{88.10} & 85.82 & \multicolumn{1}{c|}{\textbf{87.55}} & \multicolumn{1}{c|}{86.92} & 85.17 & 69.41\\ \hline
70 \%& \multicolumn{1}{c|}{\textbf{83.92}} & \multicolumn{1}{c|}{83.27}& 71.21 & \multicolumn{1}{c|}{85.92}& \multicolumn{1}{c|}{\textbf{86.39}} & 78.38 & \multicolumn{1}{c|}{\textbf{84.57}} & \multicolumn{1}{c|}{82.75} & 71.26 & 62.31\\ \hline
80 \%& \multicolumn{1}{c|}{\textbf{78.05}} & \multicolumn{1}{c|}{77.11}& 63.90 & \multicolumn{1}{c|}{\textbf{77.63}} & \multicolumn{1}{c|}{77.49} & 64.19 & \multicolumn{1}{c|}{\textbf{80.55}} & \multicolumn{1}{c|}{80.02} & 66.76 & 49.68\\ \hline \Xhline{2\arrayrulewidth}
\end{tabular}
\vspace{2mm}
\caption{Recovery results of LeNet-300-100 on FashionMNIST }
\label{tab:LeNet_FashionMNIST}
\end{table*}

\begin{table*}[h]
% \color{blue}
\centering
\small
\begin{tabular}{c||c |c |c|c|c |c |c |c |c } \Xhline{2\arrayrulewidth}
% \multicolumn{7}{c}{\textbf{ResNet-50 (Top-1 Acc. 78.82 / MACs. 64.27 G / Params. 19.68M) }} \\ \Xhline{2\arrayrulewidth} %\hline
\multicolumn{10}{c}{\textbf{ResNet-50 (Top-1 Acc. 78.82 / MACs. 1.31 G / Params. 23.71M) }} \\ \Xhline{2\arrayrulewidth} %\hline
\multicolumn{1}{c||}{Thresholds} & \multicolumn{3}{c|}{ 0.1 } & \multicolumn{3}{c|}{0.3} & \multicolumn{3}{c}{0.5} \\ \hline
 Methods & Ours & NM  &Prune & Ours & NM &Prune & Ours & NM   &Prune\\ \Xhline{2\arrayrulewidth}
% Top-1 Acc (\%) & \textbf{76.82} & 76.28 & \textbf{69.66} & 66.53 &  \textbf{50.11}& 29.4  \\ \hline
% Parmas (M) & \textbf{19.86} & 20.39& \textbf{17.47} & 19.45&  \textbf{16.47} &17.18  \\ \hline
% MACs (G) & \textbf{50.01} & 50.85&  \textbf{36.44} & 42.14&  \textbf{29.76} & 31.58  \\ \hline 
Top-1 Acc (\%) & \textbf{69.66} & 66.07 & 62.31 & \textbf{65.58} & 47.1 & 24.61 & \textbf{41.53}& 6.6 & 4.65\\ \hline
Parmas (M) & \textbf{17.47} & 20.01& 20.13 & \textbf{14.80} & 14.83& 14.85& 10.49 &10.49  & 10.49\\ \hline
MACs (G) & \textbf{0.74} & 0.90& 0.95 &  0.95 & \textbf{0.70}& 0.73& 0.53 & 0.53 & 0.53 \\ \hline 

\Xhline{2\arrayrulewidth}
\end{tabular}
\vspace{2mm}
\caption{Recovered accuracies of ResNet-50 on CIFAR100 in global adaptive pruning}
\label{tab:global_pruning_resnet50_cifar100}
\end{table*}

\begin{table}[t!]
\centering
% \color{blue}
\begin{tabular}{c||c|c|c}
\hline
      & \ MACs(G) & \ Params(M) & Accuracy(\%) \\ \hline \Xhline{2\arrayrulewidth}
Ours  & 0.74    & 17.47        & 94.05        \\ \hline
NM    & 0.90     & 20.01      & 93.88        \\ \hline
Prune & 0.95     & 20.13        & 93.98        \\ \hline
Fine-tuning  & 1.31    & 23.72        & 82.66        \\ \hline
\end{tabular}
\vspace{2mm}
\caption{Transfer-learning accuracies of ResNet-50 from CIFAR-100 to CIFAR-10}
\label{tab:downstream_cifar}
\end{table}

\begin{table}[t!]
\centering
% \color{blue}
\begin{tabular}{c||c|c|c}
\hline
      & \ MACs(G) & \ Params(M) & Accuracy(\%) \\ \hline \Xhline{2\arrayrulewidth}
Ours  & 2.69     & 15.46        & 91.98        \\ \hline
NM    & 2.84   & 17.09       & 91.36        \\ \hline
Prune & 2.96    & 17.13        & 91.8        \\ \hline
Fine-tuning  & 3.68    & 21.8       & 92.54        \\ \hline

\end{tabular}
\vspace{2mm}
\caption{Transfer-learning accuracies of ResNet-34 from ImageNet to CUB-200-2011}
\label{tab:downstream_cub200}
\end{table}

\subsection{Recovery Performance with Global Pruning}
\revsec{We examine how the recovery performance of LBYL would be if we adopt a global pruning scheme, where we adaptively changes pruning ratios across layers rather than applying the same pruning ratio to every layer. In order to measure the pruning intensity of each layer, we employ WARE between the original model and the model being pruned and recovered, and set a threshold value, which indicates the maximum WARE value during the entire pruning procedure. More specifically, we repeat an iterative procedure that first prunes a small number of filters (\textit{e.g.,} 10\%) for each layer, followed by a corresponding recovery process without training. Then, we check if the resulting WARE value exceeds a pre-defined threshold value, and stop pruning for the layer if it is the case. Varying the threshold values from 0.1 to 0.5, Table \ref{tab:global_pruning_resnet50_cifar100} demonstrates that our LBYL method outperforms NM \cite{NM} in particular for the highest pruning intensity with the threshold 0.5. Notably, it is also observed that LBYL can prune more filters and hence ends up with smaller compressed models, while still achieving higher accuracies than NM.}

\subsection{\revfin{Effectiveness of Recovering Transferable Knowledge}}

\revfin{Finally, we examine whether our method can be effective to make pruned models well-adapted to downstream tasks. To this end, we first prune and recover pretrained models on more comprehensive datasets (e.g., CIFAR-100 and ImageNet) with global pruning scheme, and then fine-tune the recovered models on less complicated (e.g., CIFAR-10) or less generic (e.g., CUB-200-2011) datasets during only 20 epochs. As clearly observed in Tables \ref{tab:downstream_cifar} and \ref{tab:downstream_cub200}, our LBYL method reaches to the highest accuracy, even with a higher compression ratio (i.e., a smaller model size). This explains the effectiveness of LBYL at recovering more transferable knowledge from a pretrained model to downstream tasks.}

%% file: conclusion.tex
\section{Conclusion} \label{sec:conclusion}
This paper proposed the problem of restoring a pruned CNN in a way free of training data and fine-tuning. We mathematically formulated how filter pruning can make a damage to the output of the pruned network, and focused on how to model the information carried by each pruned filter to be delivered to the other remaining filters. Our proposed assumption is inspired by the fact that the more the remaining filters participate in the recovery process, the better the approximation we can obtain for the original output. With this assumption, we successfully decomposed the reconstruction error into the three different components, and thereby designed a data-free loss function along with its closed form solution. Our future work would be to extend the proposed loss function so as to cover different activation functions other than ReLU.

%% file: appendix.tex
\onecolumn

\section*{Appendix \\``Training-Free Restoration of Pruned Neural Networks''} \label{sec:supp}

% \vspace{10mm}

\setcounter{table}{0}
\setcounter{figure}{0}

\renewcommand{\thetable}{A\arabic{table}}  
\renewcommand{\thefigure}{A\arabic{figure}}
\renewcommand{\thesubsection}{A\arabic{subsection}}

In this appendix, we first present the proofs of Lemma \ref{lem:bn}, Theorem \ref{thm:relu}, and Theorem \ref{thm:closedform}, respectively. Then, we describe more details about our experimental results including the information about the pretrained model and the values of hyperparameters.

\subsection{Notation Table}
Let us first present the following table of notations frequently used throughout the manuscript and Appendix.

\begin{table}[h]
\centering
\begin{tabular}{l||l}
\hline
$\mathbf{W}^{(\ell)}$ and $\Tilde{\mathbf{W}}^{(\ell)}$                 &  original and damaged (due to pruning) filters in $\ell$-th layer            \\ \hline
$\mathbf{A}^{(\ell)}$ and $\Tilde{\mathbf{A}}^{(\ell)}$         & original and damaged (due to pruning) activation maps in $\ell$-th layer                                 \\ \hline
$\mathbf{Z}^{(\ell)}$  and $\Tilde{\mathbf{Z}}^{(\ell)}$         & original and damaged (due to pruning) feature maps in $\ell$-th layer                                    \\ \hline
$\N(\cdot)$                   & batch normalization function                                     \\ \hline
$\F(\cdot)$                   & activation function such as ReLU                                             \\ \hline
$\boldsymbol{\S}$                         & pruning matrix or delivery matrix                                        \\ \hline
$\gamma, \beta, \sigma, \mu$                    & batch normalization parameters              \\ \hline
$\boldsymbol{\E}$                         & residual error (RE)                                       \\ \hline
$\boldsymbol{\B}$                         & batch normalization error (BE)                              \\ \hline
$\boldsymbol{\R}$                         & activation error (AE)                              \\ \hline
$\mathbf{s}$                              & scale coefficients for preserved filters to restore pruned filters              \\ \hline
\end{tabular}%
\vspace{2mm}
\caption{Table of notations}
\label{tab:notations}
\end{table}

\subsection{Proofs on Our Theoretical Results} \label{sec:appendix:proof}
For the ease of presentation, we assume the operators like addition `$+$', subtraction `$-$', and minimum `$\min(\cdot)$' or maximum `$\max(\cdot)$' to support \textit{broadcasting} as well as \textit{element-wise}. For example, given two equal-sized matrices $A$, $B$ and a constant $b$, $A+b$ means that every element of $A$ is added by $b$, whereas $A+B$ indicates a normal element-wise addition of two matrices.

\subsubsection{Proof of Lemma \ref{lem:bn}}
\begin{proof}
If there is only batch normalization between a feature map and its activation map, $\mathbf{A}^{(\ell)} = \N(\mathbf{Z}^{(\ell)})$. In this case, the reconstruction error can be formulated as below.
\small
\begin{eqnarray}
\begin{split}
     \|\mathbf{A}_{j}^{(\ell)}-\sum\limits_{k = 1, k \neq j}^{m}{s_{k}} \mathbf{A}_{k}^{(\ell)}\|_{1} = &~\| \N(\mathbf{Z_j^{(\ell)}}) - \sum\limits_{k = 1, k \neq j }^{m} s_{k} \N(\mathbf{Z}_k^{(\ell)}) \|_{1} \\
     =&~\| \frac{\gamma_{j}({\mathbf{Z}_j^{(\ell)} - \mu_j})}{\sigma_{j}} + \beta_j -\sum\limits_{k = 1, k \neq j}^{m}{s_{k}} \{ \frac{\gamma_{k}({\mathbf{Z}_k^{(\ell)} - \mu_k})}{\sigma_{k}} + \beta_k \} \|_{1}\\
     =&~\|\frac{\gamma_{j}}{\sigma_{j}}\mathbf{Z}_j^{(\ell)} - \sum\limits_{k = 1, k \neq j}^{m}\frac{s_{k}{\gamma_{k}}}{\sigma_{k}}\mathbf{Z}_k^{(\ell)}  -\frac{\gamma_{j}\mu_j}{\sigma_{j}} + \beta_j + \sum\limits_{k = 1, k \neq j}^{m}\{\frac{s_{k}{\gamma_{k}}\mu_k}{\sigma_{k}} - s_{k}\beta_k \}\|_{1} \\
     =&~\|\frac{\gamma_{j}}{\sigma_{j}}(\mathbf{A}^{(\ell-1)} \circledast \mathbf{W}_j^{(\ell)}) - \sum\limits_{k = 1, k \neq j}^{m}\frac{s_{k}{\gamma_{k}}}{\sigma_{k}}(\mathbf{A}^{(\ell-1)} \circledast \mathbf{W}_k^{(\ell)}) +  \\
     &~\sum\limits_{k = 1, k \neq j}^{m} \{{s_{k}} \frac{\gamma_{k}}{\sigma_{k}}(\mu_k - \frac{\sigma_{k}}{\gamma_{k}}\beta_k)\} -\frac{\gamma_{j}\mu_j}{\sigma_{j}} + \beta_j\|_{1}\\
     =&~\| \frac{\gamma_{j}}{\sigma_{j}} \{ (\mathbf{A}^{(\ell-1)} \circledast \mathbf{W}_j^{(\ell)})  - \sum\limits_{k = 1, k \neq j}^{m}s_k\frac{\sigma_{j}\gamma_{k}}{\gamma_{j}\sigma_{k}}(\mathbf{A}^{(\ell-1)} \circledast \mathbf{W}_k^{(\ell)}) \}  + \\
     &~\frac{\gamma_{j}}{\sigma_{j}}\sum\limits_{k = 1, k \neq j}^{m} \{s_k \frac{\sigma_{j}\gamma_{k}}{\gamma_{j}\sigma_{k}}(\mu_k - \frac{\sigma_{k}}{\gamma_{k}}\beta_k) - \mu_j +\frac{\sigma_{j}}{\gamma_{j}}\beta_j \}\|_{1}\\
     =&~\|\frac{\gamma_{j}}{\sigma_{j}} \{\mathbf{A}^{(\ell-1)} \circledast (\mathbf{W}_j^{(\ell)} - \sum\limits_{k = 1, k \neq j}^{m}s_k\frac{\sigma_{j}\gamma_{k}}{\gamma_{j}\sigma_{k}}\mathbf{W}_k^{(\ell)})\} \\
     &~+ \frac{\gamma_{j}}{\sigma_{j}}\sum\limits_{k = 1, k \neq j}^{m} \{s_k \frac{\sigma_{j}\gamma_{k}}{\gamma_{j}\sigma_{k}}(\mu_k - \frac{\sigma_{k}}{\gamma_{k}}\beta_k) - \mu_j +\frac{\sigma_{j}}{\gamma_{j}}\beta_j \} \|_{1}\\
     =&~\|\frac{\gamma_{j}}{\sigma_{j}}(\mathbf{A}^{(\ell-1)} \circledast \boldsymbol{\E}) + \frac{\gamma_{j}}{\sigma_{j}}\sum\limits_{k = 1, k \neq j}^{m} \{s_k \frac{\sigma_{j}\gamma_{k}}{\gamma_{j}\sigma_{k}}(\mu_k - \frac{\sigma_{k}}{\gamma_{k}}\beta_k) - \mu_j +\frac{\sigma_{j}}{\gamma_{j}}\beta_j \} \|_{1},
     \nonumber
\end{split}
\end{eqnarray}
where $\boldsymbol{\E}$ = $\mathbf{W}_j^{(\ell)} - \sum\limits_{k = 1, k \neq j}^{m}s_k\frac{\sigma_{j}\gamma_{k}}{\gamma_{j}\sigma_{k}}\mathbf{W}_k^{(\ell)} $.
\end{proof}

\subsubsection{Proof of Theorem \ref{thm:relu}}

\begin{proof}
If there are both batch normalization and a ReLU function between a feature map and its activation map, $\mathbf{A}^{(\ell)} = \F(\N(\mathbf{Z}^{(\ell)}))$. In this case, the reconstruction error can be formulated as below.
% \scriptsize
\small
\begin{eqnarray}
\begin{split}
 \|\mathbf{A}_{j}^{(\ell)}-\sum\limits_{k = 1, k \neq j}^{m}{s_{k}} \mathbf{A}_{k}^{(\ell)}\|_{1} &=\|\max(\N(\mathbf{Z}_{j}^{(\ell)}),0) - \sum\limits_{k=1, k \neq j}^{m} s_{k} \max(\N(\mathbf{Z}_{k}^{(\ell)}),0)\|_{1}\\
  &=\|\N(\mathbf{Z}_{j}^{(\ell)}) - \sum\limits_{k = 1, k \neq j}^{m}{s_{k}}\N(\mathbf{Z}_k^{(\ell)}) + \sum\limits_{k=1, k \neq j}^{m} s_{k} \min(0,\N(\mathbf{Z}_{k}^{(\ell)})) - \min(0, \N(\mathbf{Z}_j^{(\ell)}))\|_{1}\\
\end{split}\nonumber
\end{eqnarray}
From the procedure of Proof of Lemma \ref{lem:bn}, we obtained $\N(\mathbf{Z}_j^{(\ell)}) - \sum\limits_{k = 1, k \neq j }^{m} s_{k} \N(\mathbf{Z}_k^{(\ell)}) $ = $\frac{\gamma_{j}}{\sigma_{j}}(\mathbf{A}^{(\ell-1)} \circledast \boldsymbol{\E}) + \boldsymbol{\B}$, where $\boldsymbol{\E}$ = $\mathbf{W}_j^{(\ell)} - \sum\limits_{k = 1, k \neq j}^{m}s_k\frac{\sigma_{j}\gamma_{k}}{\gamma_{j}\sigma_{k}}\mathbf{W}_k^{(\ell)} $ and $\boldsymbol{\B} =\frac{\gamma_{j}}{\sigma_{j}}\sum\limits_{k = 1, k \neq j}^{m} \{s_k \frac{\sigma_{j}\gamma_{k}}{\gamma_{j}\sigma_{k}}(\mu_k - \frac{\sigma_{k}}{\gamma_{k}}\beta_k) - \mu_j +\frac{\sigma_{j}}{\gamma_{j}}\beta_j \}$.
Therefore, we have:
\begin{equation}
    \|\mathbf{A}_{j}^{(\ell)}-\sum\limits_{k = 1, k \neq j}^{m}{s_{k}} \mathbf{A}_{k}^{(\ell)}\|_{1} = \|\frac{\gamma_{j}}{\sigma_{j}}(\mathbf{A}^{(\ell-1)} \circledast \boldsymbol{\E}) + \boldsymbol{\B} + \boldsymbol{\R} \|_{1},
    \nonumber
\end{equation}
where $\boldsymbol{\R} = \sum\limits_{k=1, k \neq j}^{m} s_{k}  \min(0,\N(\mathbf{Z}_{k}^{(\ell)}))-\min(0,\N(\mathbf{Z}_{j}^{(\ell)}))$.
\end{proof}

\subsubsection{\revsec{Proof of Lemma \ref{lem:ae}}}
\revsec{
\begin{proof}
\begin{eqnarray}
    \|\boldsymbol{\R}\|_{1} & =  & \|\sum_{k=1, k \neq j}^{m} s_{k}  \min(0,\N(\mathbf{Z}_{k}^{(\ell)}))-\min(0,\N(\mathbf{Z}_{j}^{(\ell)})) \|_{1} \nonumber \\
    & =  & \|\sum_{k=1, k \neq j}^{m} s_{k}  \min(0,\N(\mathbf{Z}_{k}^{(\ell)}))\|_{1}-\min(0,\N(\mathbf{Z}_{j}^{(\ell)})) \nonumber \\
    & \leq & \sum_{k=1,k\neq j}^{m} \| s_k \min(0,\N(\mathbf{Z}_{k}^{(\ell)}))\|_{1} -\min(0,\N(\mathbf{Z}_{j}^{(\ell)})) \nonumber \\
    & \leq & \sum_{k=1,k\neq j}^{m}\|s_k \cdot \N(\mathbf{Z}_{k}^{(\ell)})\|_{1} -\min(0,\N(\mathbf{Z}_{j}^{(\ell)})) \nonumber \\
    & = & \sum_{k=1,k\neq j}^{m}\|s_k\|_{1} \cdot \|\N(\mathbf{Z}_{k}^{(\ell)})\|_{1} +c, \nonumber 
\end{eqnarray}
where $c = -\min(0,\N(\mathbf{Z}_{j}^{(\ell)})) \geq 0$.
\end{proof}
}

\subsubsection{Proof of Theorem \ref{thm:closedform}}
\begin{proof}
Our loss function is as follows.
\small
\begin{equation}
     \mathcal{L}_{re} =  \|\boldsymbol{\E}\|_2^{2} + \lambda_1 \|\boldsymbol{\B}\|_2^{2} + \lambda_2 \|\mathbf{s}\|_{2}^{2},
\nonumber
\end{equation}
where $\boldsymbol{\E}$ = $\mathbf{W}_j^{(\ell)} - \sum\limits_{k = 1, k \neq j}^{m}s_k\frac{\sigma_{j}\gamma_{k}}{\gamma_{j}\sigma_{k}}\mathbf{W}_k^{(\ell)} $,~$\boldsymbol{\B} =\frac{\gamma_{j}}{\sigma_{j}}\sum\limits_{k = 1, k \neq j}^{m} \{s_k \frac{\sigma_{j}\gamma_{k}}{\gamma_{j}\sigma_{k}}(\mu_k - \frac{\sigma_{k}}{\gamma_{k}}\beta_k) - \mu_j +\frac{\sigma_{j}}{\gamma_{j}}\beta_j \},~\mathbf{s} =  [s_{1}~...~s_{j-1}~s_{j+1}~...~s_{m}]^T,~\lambda_1 , \lambda_2 > 0$. \\
\\
\\
Let (1) $\mathbf{X}=[\frac{\sigma_{j}\gamma_{1}}{\gamma_{j}\sigma_{1}}\boldsymbol{f_1},...,\frac{\sigma_{j}\gamma_{j-1}}{\gamma_{j}\sigma_{j-1}}\boldsymbol{f_{j-1}},\frac{\sigma_{j}\gamma_{j+1}}{\gamma_{j}\sigma_{j+1}}\boldsymbol{f_{j+1}},...\frac{\sigma_{j}\gamma_{m}}{\gamma_{j}\sigma_{m}}\boldsymbol{f_m}]$ such that $\boldsymbol{f_i}$ is the vectorized  $\mathbf{W}_i^{(\ell)}$ for $i \in [1,m] \setminus \{j\}$, (2) $\mathbf{y}$ is the vectorized $\mathbf{W}_j^{(\ell)}$ and (3) $\mathbf{p}=[p_1,...,p_{j-1}, p_{j+1}, ... p_n]^{T}$ such that $p_i = \frac{\sigma_{j}\gamma_{i}}{\gamma_{j}\sigma_{i}}(\mu_i - \frac{\sigma_{i}}{\gamma_{i}}\beta_i)$.\\
Then, Our loss function can be represented as below.
\begin{equation}
 \mathcal{L}_{re}(\mathbf{s}) = (\mathbf{y} - \mathbf{X}\mathbf{s})^{T}(\mathbf{y} - \mathbf{X}\mathbf{s}) + \lambda_1\{\frac{\gamma_{j}}{\sigma_{j}}(\mathbf{s}^{T}\mathbf{p} - \mu_j + \frac{\sigma_{j}}{\gamma_{j}}\beta_j)\}^2 + \lambda_2\mathbf{s}^T\mathbf{s}.
    \nonumber
\end{equation}
The first derivative of the loss function is: 
\begin{equation}
\begin{split}
\frac{\partial\mathcal{L}_{re}(\mathbf{s})}{\partial\mathbf{s}} &=  -2\mathbf{X}^T\mathbf{y} + 2\mathbf{X}^T\mathbf{X}\mathbf{s} + \lambda_1 \{\frac{\gamma_{j}^2}{\sigma_{j}^2}(2\mathbf{p}\mathbf{p}^T)\mathbf{s} - 2 \frac{\gamma_{j}^2}{\sigma_{j}^2}\mu_j\mathbf{p} + 2\frac{\gamma_{j}}{\sigma_{j}}\beta_j\mathbf{p}\} + \lambda_2(2\mathbf{s}) \\
&= [2X^{T}X+2\lambda_{1}\frac{\gamma_{j}^2}{\sigma_{j}^2}\boldsymbol{p}\boldsymbol{p}^{T}+2\lambda_{2}I]\mathbf{s} -2 X^{T}\boldsymbol{y}-2\frac{\lambda_{1}\gamma_{j}}{\sigma_{j}}(\frac{\mu_{j}\gamma_{j}}{\sigma_{j}}-\beta_{j})\boldsymbol{p}
\end{split}
 \nonumber
\end{equation}

The second derivative of the loss function(\textit{i.e.} Hessian) is: 
\begin{equation}
 \mathbf{H}_{\mathcal{L}_{re}} = \frac{\partial^2\mathcal{L}_{re}(\mathbf{s})}{\partial\mathbf{s}\partial\mathbf{s}^T} = 2\mathbf{X}^T\mathbf{X} + 2\lambda_1\frac{\gamma_{j}^2}{\sigma_{j}^2}(\mathbf{p}\mathbf{p}^T)+2\lambda_2I.
 \nonumber
\end{equation}
Since $\mathbf{H}_{\mathcal{L}_{re}}$ is positive definite, $\mathcal{L}_{re}$ is a convex function and thus there exists a unique optimal solution $\mathbf{s}$ such that $\frac{\partial\mathcal{L}_{re}(\mathbf{s})}{\partial\mathbf{s}} = \mathbf{0}$. From the first derivative above, we obtain the solution as below.

\begin{equation}
 \boldsymbol{s}=[X^{T}X+\lambda_{1}\frac{\gamma_{j}^2}{\sigma_{j}^2}\boldsymbol{p}\boldsymbol{p}^{T}+\lambda_{2}I]^{-1}[X^{T}\boldsymbol{y}+\frac{\lambda_{1}\gamma_{j}}{\sigma_{j}}(\frac{\mu_{j}\gamma_{j}}{\sigma_{j}}-\beta_{j})\boldsymbol{p}].
\nonumber
\end{equation}
\end{proof}

\subsection{Experimental Details} \label{sec:appendix:exp}

\smalltitle{Hyperparameters}
LBYL uses two hyperparameters, namely $\lambda_1$ and $\lambda_2$, which adjust the weights of loss terms BE and AE, respectively. NM \cite{NM} also needs two hyperparameters, namely $\lambda$ and $t$, where $\lambda$ is the ratio for balancing between the cosine distance and the bias distance and $t$ is the threshold for avoiding erroneous compensation caused by the low similarity. In \cite{NM}, $\lambda$ is 0.85 and $t$ is 0.1 by default, which we also adopt in our experiments. However, these values do not work well in random pruning and therefore we had to find the best values of $\lambda$ and $t$ as well as $\lambda_1$ and $\lambda_2$. All hyperparameter values are found by grid search, and presented in Tables \ref{tab:param:vgg16:cifar10}, \ref{tab:param:ResNet50:cifar100}, \ref{tab:param:ResNet34:ImageNet}, and \ref{tab:param:ResNet101:ResNet101}.

\begin{table*}[h]
\centering 
\scriptsize
\begin{tabular}{c|c|c|c}\Xhline{2\arrayrulewidth}
\multirow{2}{*}{Criterion} & L2-norm& L2-GM & L1-norm\\ \cline{2-4} 
& OURS& OURS& OURS\\ \hline
Pruning Ratio& $\lambda$  & $\lambda$    & $\lambda$ \\\Xhline{2\arrayrulewidth}
50\%  & $0.3$  & $1.2$ & $0.7$  \\ \hline
60\%  & $0.6$  & $0.5$ & $0.8$  \\ \hline
70\%  & $0.3$  & $1.3$ & $0.2$   \\ \hline
80\%  & $1\times10^{-6}$  & $0.003$ & $0.3$  \\ \hline\Xhline{2\arrayrulewidth}
\end{tabular}% 
\caption{hyperparameters of LeNet-300-100 on FashionMNIST}
\label{tab:param:LeNet:fashionmnist}
\vspace{4mm}
% \end{table*}

% \begin{table*}[h]
\centering 
\scriptsize
\begin{tabular}{c|c|c|c}\Xhline{2\arrayrulewidth}
\multirow{2}{*}{Criterion} & L2-norm& L2-GM & L1-norm\\ \cline{2-4} 
& OURS& OURS& OURS\\ \hline
Pruning Ratio& $\lambda_1$ / $\lambda_2$  & $\lambda_1$ / $\lambda_2$  & $\lambda_1$ / $\lambda_2$  \\\Xhline{2\arrayrulewidth}
10\%  & $6\times10^{-6}$ / $1\times10^{-4}$ & $1\times10^{-6}$ / $4\times10^{-3}$ & $2\times10^{-6}$ / $0.06$  \\ \hline
20\%  & $4\times10^{-6}$ / $6\times10^{-3}$ & $4\times10^{-6}$ / $4\times10^{-3}$ & $2\times10^{-6}$ / $1\times10^{-4}$ \\ \hline
30\%  & $1\times10^{-6}$ / $0.01$           & $1\times10^{-6}$ / $5\times10^{-3}$ & $4\times10^{-6}$ / $1\times10^{-4}$ \\ \hline
40\%  & $2\times10^{-6}$ / $0.01$           & $1\times10^{-6}$ / $8\times10^{-3}$ & $1\times10^{-6}$ / $1\times10^{-4}$ \\ \hline
50\%  & $4\times10^{-5}$ / $2\times10^{-4}$ & $2\times10^{-5}$ / $1\times10^{-4}$ & $1\times10^{-6}$ / $0.01$  \\ \Xhline{2\arrayrulewidth}
\end{tabular}% 
\vspace{4mm}
% \end{table*}
\begin{tabular}{c|c|c|c|c|c|c}\Xhline{2\arrayrulewidth}
\multirow{2}{*}{Criterion} & \multicolumn{2}{c|}{Random\_1}& \multicolumn{2}{c|}{Random\_2}& \multicolumn{2}{c}{Random\_3}\\ \cline{2-7} 
    & OURS& NM& OURS& NM& OURS& NM\\ \hline
Pruning Ratio& $\lambda_1$ / $\lambda_2$ & t / $\lambda$ & $\lambda_1$ / $\lambda_2$ & t / $\lambda$ & $\lambda_1$ / $\lambda_2$ & t / $\lambda$ \\ \Xhline{2\arrayrulewidth}
10\%& 0.9 / $4\times10^{-4}$& 0.2 / 0.7   & $5\times10^{-6}$ / 0.2& 0.25 / 0.65     & $1\times10^{-6}$ / 0.2  & 0.05 / 0.8\\ \hline
20\%& $1\times10^{-5}$ / 0.6& 0 / 0.75    & $6\times10^{-6}$ / 0.2& 0.15 / 0.6& $2\times10^{-6}$ / 0.08 & 0.35 / 0.05 \\ \hline
30\%& $1\times10^{-5}$ / $2\times10^{-4}$ & 0 / 0.75    & $6\times10^{-6}$ / 0.09& 0.15/ 0.75& $2\times10^{-6}$ / $9\times10^{-3}$   & 0.05 / 1.0  \\ \hline
40\%& $1\times10^{-5}$ / $9\times10^{-3}$ & 0.15 / 0.7  & $3\times10^{-6}$ / 0.02& 0.05/ 1.0 & $2\times10^{-6}$ / $9\times10^{-3}$   & 0.15 / 1.0  \\ \hline
50\%& $1\times10^{-5}$ / $3\times10^{-3}$ & 0.15 / 0.75 & $1\times10^{-5}$ /~$7\times10^{-5}$ & 0.45 / 0.95     & $4\times10^{-6}$ / $7\times10^{-5}$   & 0.15 / 1.0  \\ \Xhline{2\arrayrulewidth}
\end{tabular}% 
\caption{hyperparameters of VGG16 on CIFAR-10}
\label{tab:param:vgg16:cifar10}
\vspace{4mm}
% \end{table*}
% \begin{table*}[h]
\centering 
\scriptsize
\begin{tabular}{c|c|c|c}\Xhline{2\arrayrulewidth}
\multirow{2}{*}{Criterion} & L2-norm& L2-GM & L1-norm\\ \cline{2-4} 
& OURS& OURS& OURS\\ \hline
Pruning Ratio& $\lambda_1$ / $\lambda_2$  & $\lambda_1$ / $\lambda_2$  & $\lambda_1$ / $\lambda_2$  \\\Xhline{2\arrayrulewidth}
10\%  & $2\times10^{-5}$ / $6\times10^{-3}$ & $2\times10^{-5}$ / $1\times10^{-3}$ & $2\times10^{-5}$ / $1\times10^{-3}$ \\ \hline
20\%  & $1\times10^{-5}$ / $2\times10^{-3}$ & $1\times10^{-5}$ / $1\times10^{-3}$ & $2\times10^{-5}$ / $1\times10^{-3}$ \\ \hline
30\%  & $1\times10^{-5}$ / $2\times10^{-3}$ & $1\times10^{-5}$ / $2\times10^{-3}$ & $1\times10^{-5}$ / $1\times10^{-3}$ \\ \hline
40\%  & $1\times10^{-5}$ / $1\times10^{-3}$ & $1\times10^{-5}$ / $1\times10^{-3}$ & $1\times10^{-5}$ / $1\times10^{-3}$ \\ \hline
50\%  & $1\times10^{-6}$ / $1\times10^{-3}$ & $1\times10^{-6}$ / $1\times10^{-4}$ & $1\times10^{-6}$ / $1\times10^{-3}$  \\ \Xhline{2\arrayrulewidth}
\end{tabular}% 
\vspace{1mm}

\begin{tabular}{c|c|c|c|c|c|c}\Xhline{2\arrayrulewidth}
\multirow{2}{*}{Criterion} & \multicolumn{2}{c|}{Random\_1}& \multicolumn{2}{c|}{Random\_2}& \multicolumn{2}{c}{Random\_3}\\ \cline{2-7} 
    & OURS& NM& OURS& NM& OURS& NM\\ \hline
Pruning Ratio& $\lambda_1$ / $\lambda_2$ & t / $\lambda$ & $\lambda_1$ / $\lambda_2$ & t / $\lambda$ & $\lambda_1$ / $\lambda_2$ & t / $\lambda$ \\ \Xhline{2\arrayrulewidth}
10\%& $1\times10^{-5}$ / $2\times10^{-3}$ & 0.2 / 0.5    & $2\times10^{-5}$ / $2\times10^{-3}$   & 0.15 / 0.65  & $3\times10^{-6}$ / $2\times10^{-3}$   & 0.2 / 0.95 \\ \hline
20\%& $1\times10^{-5}$ / $3\times10^{-3}$ & 0.2 / 0.5    & $4\times10^{-5}$ / $1\times10^{-4}$   & 0.05 / 0.7   & $4\times10^{-6}$ / $5\times10^{-3}$   & 0.3 / 0.55 \\ \hline
30\%& $1\times10^{-5}$ / $2\times10^{-3}$ & 0.4 / 0.6    & $3\times10^{-6}$ / $2\times10^{-3}$   & 0.15/ 0.7    & $1\times10^{-6}$ / $4\times10^{-3}$   & 0.05 / 0.9  \\ \hline
40\%& $1\times10^{-5}$ / $2\times10^{-3}$ & 0.2 / 0.5    & $8\times10^{-6}$ / $7\times10^{-4}$   & 0.4 / 0.9     & $1\times10^{-6}$ / $4\times10^{-3}$   & 0.25 / 0.55  \\ \hline
50\%& $1\times10^{-5}$ / $2\times10^{-3}$ & 0.4 / 1.0    & $6\times10^{-6}$ /~$9\times10^{-4}$   & 0.1 / 0.7    & $1\times10^{-6}$ / $5\times10^{-3}$   & 0.05 / 1.0  \\ \Xhline{2\arrayrulewidth}
\end{tabular}% 
\caption{hyperparameters of ResNet50 on CIFAR-100}
\label{tab:param:ResNet50:cifar100}
\vspace{2mm}
\end{table*}

\begin{table*}[h]
\centering 
\scriptsize
\begin{tabular}{c|c|c|c}\Xhline{2\arrayrulewidth}
\multirow{2}{*}{Criterion} & L2-norm& L2-GM & L1-norm\\ \cline{2-4} 
& OURS& OURS& OURS\\ \hline
Pruning Ratio& $\lambda_1$ / $\lambda_2$    & $\lambda_1$ / $\lambda_2$  & $\lambda_1$ / $\lambda_2$  \\\Xhline{2\arrayrulewidth}
10\%  & $7\times10^{-5}$ / 0.05 & $1\times10^{-5}$ / 0.1 &  $3\times10^{-5}$ / 0.08 \\ \hline
20\%  & $2\times10^{-5}$ / 0.07 & $2\times10^{-4}$ / 0.06 & $5\times10^{-4}$ / 0.03 \\ \hline
30\%  & $5\times10^{-4}$ / 0.03 & $4\times10^{-4}$ / 0.04 & $8\times10^{-5}$ / 0.05 \\ \Xhline{2\arrayrulewidth}
\end{tabular}% 
\vspace{1mm}

\begin{tabular}{c|c|c|c|c|c|c}\Xhline{2\arrayrulewidth}
\multirow{2}{*}{Criterion} & \multicolumn{2}{c|}{Random\_1}& \multicolumn{2}{c|}{Random\_2}& \multicolumn{2}{c}{Random\_3}\\ \cline{2-7} 
    & OURS& NM& OURS& NM& OURS& NM\\ \hline
Pruning Ratio& $\lambda_1$ / $\lambda_2$ & t / $\lambda$ & $\lambda_1$ / $\lambda_2$ & t / $\lambda$ & $\lambda_1$ / $\lambda_2$ & t / $\lambda$ \\ \Xhline{2\arrayrulewidth}
10\%& $4\times10^{-4}$ / 0.07 & 0.15 / 0.1    & $1\times10^{-5}$ / 0.3   & 0.05 / 0.75   & $1\times10^{-5}$ / 0.08   & 0.1 / 0.6 \\ \hline
20\%& $1\times10^{-5}$ / 0.1  & 0.15 / 1.0    & $1\times10^{-5}$ / 0.3   & 0.05 / 0.65   & $1\times10^{-5}$ / 0.2    & 0.1 / 0.8 \\ \hline
30\%& $2\times10^{-5}$ / 0.2  & 0.15 / 0.9    & $4\times10^{-5}$ / 0.2   & 0.05 / 0.65   & $2\times10^{-5}$ / 0.07   & 0.05 / 0.85  \\ \Xhline{2\arrayrulewidth}
\end{tabular}% 
\caption{hyperparameters of ResNet34 on ImageNet}
\label{tab:param:ResNet34:ImageNet}
\vspace{2mm}

% \end{table*}

% \begin{table*}[h]
\centering 
\scriptsize
\begin{tabular}{c|c|c|c}\Xhline{2\arrayrulewidth}
\multirow{2}{*}{Criterion} & L2-norm& L2-GM & L1-norm\\ \cline{2-4} 
& OURS& OURS& OURS\\ \hline
Pruning Ratio& $\lambda_1$ / $\lambda_2$    & $\lambda_1$ / $\lambda_2$  & $\lambda_1$ / $\lambda_2$  \\\Xhline{2\arrayrulewidth}
10\%  & $4\times10^{-6}$ / 0.02 & $8\times10^{-6}$ / $9\times10^{-4}$   &  $1\times10^{-6}$ / $5\times10^{-3}$ \\ \hline
20\%  & $1\times10^{-6}$ / 0.02 & $1\times10^{-6}$ / 0.01               & $1\times10^{-6}$ / 0.02 \\ \hline
30\%  & $2\times10^{-6}$ / 0.03 & $1\times10^{-6}$ / 0.02               & $1\times10^{-6}$ / 0.03 \\ \Xhline{2\arrayrulewidth}
\end{tabular}% 
\vspace{1mm}
\begin{tabular}{c|c|c|c|c|c|c}\Xhline{2\arrayrulewidth}
\multirow{2}{*}{Criterion} & \multicolumn{2}{c|}{Random\_1}& \multicolumn{2}{c|}{Random\_2}& \multicolumn{2}{c}{Random\_3}\\ \cline{2-7} 
    & OURS& NM& OURS& NM& OURS& NM\\ \hline
Pruning Ratio& $\lambda_1$ / $\lambda_2$ & t / $\lambda$ & $\lambda_1$ / $\lambda_2$ & t / $\lambda$ & $\lambda_1$ / $\lambda_2$ & t / $\lambda$ \\ \Xhline{2\arrayrulewidth}
10\%& $1\times10^{-6}$ / 0.2 & 0.1 / 1.0      & $2\times10^{-6}$ / 1.0   & 0.15 / 0.35   & $2\times10^{-6}$ / 0.06   & 0.35 / 0.9 \\ \hline
20\%& $1\times10^{-6}$ / 0.3  & 0.15 / 0.1    & $4\times10^{-5}$ / 0.2   & 0.15 / 0.45   & $6\times10^{-6}$ / 0.05    & 0.35 / 1.0 \\ \hline
30\%& $1\times10^{-6}$ / 0.2  & 0.5 / 0.85    & $1\times10^{-6}$ / 0.2   & 0.25 / 0.3   & $2\times10^{-6}$ / $1\times10^{-4}$   & 0.1 / 0.1 \\ \Xhline{2\arrayrulewidth}
\end{tabular}% 
\caption{hyperparameters of ResNet101 on ImageNet}
\label{tab:param:ResNet101:ResNet101}
\end{table*}